\documentclass[final]{cvpr}

\usepackage{graphicx}
\usepackage{amsmath}
\usepackage{amssymb}
\usepackage[mathscr]{euscript}  
\usepackage{booktabs}
\usepackage{enumitem}
\usepackage{cuted}
\usepackage{flushend}
\usepackage{times}
\usepackage{siunitx}
\usepackage{adjustbox}
\usepackage{array}
\usepackage{multirow}
\usepackage{multicol}
\usepackage{microtype}
\usepackage[resetlabels]{multibib}
\usepackage{cuted}
\usepackage[font={footnotesize}]{caption}
\usepackage{bm}
\usepackage{url}

\usepackage{subcaption}
\usepackage[acronym,smallcaps]{glossaries}
\newacronym{rpn}{RPN}{Region Proposal Network}
\newacronym{roi}{RoI}{Region of Interest}
\newacronym{iou}{IoU}{intersection over union}
\newacronym{nms}{NMS}{non-maximum suppression}
\newacronym{rgcn}{R-GCNs}{Relational Graph Convolutional Networks \cite{Schlichtkrull2018}}
\newacronym{fpn}{FPN}{Feature Proposal Network}
\newacronym{kern}{KERN}{Knowledge-Embedded Routing Network \cite{Chen2019}}
\newacronym{bert}{BERT}{Bidirectional Encoder Representations from Transformers \cite{Devlin2019}}
\newacronym{gru}{GRU}{Gated Recurrent Cells \cite{Li2016}}
\newacronym{fcn}{FCN}{fully-connected network}
\newacronym{rcgn}{RCGN}{Relational Graph Neural Network}
\newacronym{ggnn}{GGNN}{Gated-Graph Neural Network}
\newacronym{mlp}{MLP}{Multi-Layer Perceptron}
\newacronym{coco}{COCO 2017}{Common Objects in Context \cite{Lin2014}}
\newacronym{svg}{SVG}{singular value decomposition}
\newacronym{kge}{KGE}{knowledge graph embedded}

\usepackage{makecell}

\usepackage{xcolor}
\definecolor{dark-green}{RGB}{12,80,12}

\usepackage[pagebackref,breaklinks,colorlinks]{hyperref}
\usepackage{siunitx}
\usepackage[ruled,vlined]{algorithm2e}
\usepackage{algorithmic}
\usepackage{microtype}
\SetAlFnt{\small}
\SetAlCapFnt{\small}
\SetAlCapNameFnt{\small}

\usepackage{caption}
\usepackage[capitalize]{cleveref}
\crefname{section}{Sec.}{Secs.}
\Crefname{section}{Section}{Sections}
\Crefname{table}{Table}{Tables}
\crefname{table}{Tab.}{Tabs.}
\crefname{algorithm}{Algo.}{Algos.}

\def\cvprPaperID{10987} % *** Enter the CVPR Paper ID here

\def\confYear{2022}

\begin{document}

\title{Contrastive Object Detection Using Knowledge Graph Embeddings}

\author{Christopher Lang$^{1,2}$ \hspace{1cm} Alexander Braun$^{1}$ \hspace{1cm} Abhinav Valada$^{2}$
\\
\centerline{$^{1}$Robert Bosch GmbH \hspace{1cm} $^{2}$University of Freiburg}
\\
\centerline{{\tt\small christopher.lang@de.bosch.com \hspace{1cm} valada@cs.uni-freiburg.de}}
}
\maketitle

%%%%%%%%% ABSTRACT
\begin{abstract}
Object recognition for the most part has been approached as a one-hot problem that treats classes to be discrete and unrelated. 
Each image region has to be assigned to one member of a set of objects, including a background class, disregarding any similarities in the object types. 
In this work, we compare the error statistics of the class embeddings learned from a one-hot approach with semantically structured embeddings from natural language processing or knowledge graphs that are widely applied in open world object detection. 
Extensive experimental results on multiple knowledge-embeddings as well as distance metrics indicate that knowledge-based class representations result in more semantically grounded misclassifications while performing on par compared to one-hot methods on the challenging COCO and Cityscapes object detection benchmarks.
We generalize our findings to multiple object detection architectures by proposing a knowledge-embedded design \acrfull{kge} for keypoint-based and transformer-based object detection architectures. 
\end{abstract}

%%%%%%%%% BODY TEXT
\section{Introduction}
\label{sec:intro}
% Human drivers are thought of modelling causal interactions between models to make sense of a scene.
% Common object detectors on the other hand are trained on a statistical relationships of objects on single frames.
Object detection is the task of localizing and classifying objects in an image.
Deep object detection originates from the two-stage approach~\cite{ren2015:faster-rcnn, Cai2018, valverde2021there} which first searches for anchor points in the image that potentially contain objects, followed by classifying and regressing the coordinates for each proposal region in isolation.
The Faster~R-CNN approach~\cite{ren2015:faster-rcnn} achieves an average precision of 21.9\%  on the challenging \acrshort{coco} benchmark~\cite{Lin2014} for object detection in everyday situations. 
Its dependence on a heuristic suppression of multiple detections for the same object is a commonly addressed design flaw~\cite{Liu2019c}.
In an effort to make the network end-to-end trainable, recent architectures model the relationship between object proposals by formulating object detection as a keypoint detection~\cite{Zhou2019b, Law2020} or a direct-set estimation problem~\cite{Carion2020, Zhu2020, dai2021dynamic}. 
Such novel architectures along with stronger backbones~\cite{he2016deep,sirohi2021efficientlps} and multiscale feature learning techniques~\cite{lin2017feature, Qiao2020, gosala2021bird} have doubled the average precision on the COCO detection benchmark to 54.0\% \cite{dai2021dynamic}.

\begin{figure}%
    \footnotesize
	\centering
	% \includegraphics[width=\columnwidth]{graphics/glove_embeddings}
	% \caption{TNSE projection of object type embeddings used in the contrastive loss formulation. The embeddings show the GloVe features for the object type names trained on the Wikimedia 2017 dataset.}
	\includegraphics[width=0.8\linewidth]{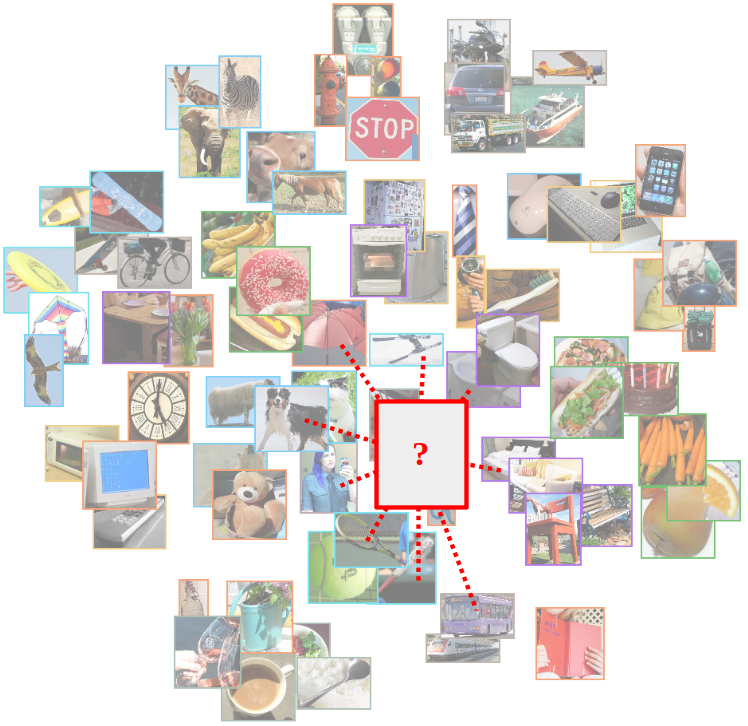}
	\caption{We incorporate class prototypes such as GloVE \cite{glove} from natural language processing, into the prediction head of common object detectors. The resulting feature space is structured by the context from non-visual knowledge sources such that similar concepts are in proximity, which we show yields more semantically-grounded predictions.}
	\label{fig:cover_example}
	\vspace{-0.2cm}
\end{figure}

These astonishing benchmark performances have grown at the expense of a set-based classification formulation, i.e., the classes are regarded as discrete and unrelated.
This assumption allows such methods to use a multinomial logistic loss~\cite{ren2015:faster-rcnn, Sun2020, Zhou2019b} 
% and a linear classification layer with the output size equal to the number of classes, 
to boost the classification accuracy by maximizing the inter-class distances~\cite{Kornblith2020}. However, this forces the model to treat each class prototype out-of-context from what it has learned about the appearance of the other classes. As a result, the training is prone to inherit bias through class-imbalance~\cite{Lin2020, He2017} or co-occurrence statistics~\cite{singh2020don} from the training data, and neglects synergies in the real world, where there exist semantic similarities between classes.
This closed-set formulation further requires drastic changes of the class prototypes to be learned whenever the set of classes is extended~\cite{Shmelkov2017, perez2020incremental}, resulting in catastrophic forgetting of previously learned classes when the training is performed incrementally~\cite{perez2020incremental, zurn2020self}.

In zero-shot learning, that aims to detect classes that were excluded from the training set, the parametric class prototypes are replaced with a mapping from visual to a semantic embedding space, e.g. word embeddings generated from text corpora, which describe semantic context for a broad set of object classes.
We extend this approach to keypoint-based and transformer-based object detectors and evaluate various knowledge embeddings as well as loss functions on the \acrshort{coco}~\cite{Lin2014} and Cityscapes~\cite{Cordts2016} datasets.
Additionally, we analyze the classification error distributions of our proposed \acrshort{kge} architectures with baseline methods of set-based classification and show that \acrshort{kge} predictions tend to be more semantically consistent with the groundtruth.
% , as can be seen in \tabref{tab:coco_confusions} for selected classes of the \acrshort{coco} dataset. 

% While existing methods focus on utilizing visual features only, we rely on the vast amount of commonsense expertise already existing in other knowledge domains, by using contextual embeddings from natural language processing or knowledge graphs. % Our approach

% the regression target vectors can distributed to reflect object hierarchies.
% The latter would allow for hierarchical object detection, for example could all vehicle object types, (like car, bus, truck, etc.) populate a concave region of the feature space.
% By replacing all vehicle object types with a single vehicle target vector located at the center of this concave region, we could further design a decision-tree like object categorization method.

% Tease your idea with few sentences and a "pull" figure

% Summarize in 3 bullets your contributions
In summary, our main contributions in this work are:
\begin{itemize}[topsep=0pt,noitemsep]
    \item a knowledge embedding formulation for two-stage, keypoint-based and transformer-based multi-object detection architectures.
    \item a quantitative ablation study on the \acrshort{coco} over different distance metrics and knowledge types.
    \item an error overlap analysis between one-hot and embedding-based classifications.
\end{itemize}

\section{Related Work}
\label{sec:relatedwork}
% Theme: how is the classification in the recognition part performed, where is it used
% How is object detection evolved?
% How was the classifciation modified
This work relates to the research areas of object detection architectures, especially their classification losses and the usage of commonsense knowledge in computer vision. In this section, we review the relevant works in these areas.

{\parskip=5pt
\noindent\textit{Object Detection}: Two-stage methods~\cite{ren2015:faster-rcnn, Cai2018, Sun2020} rely on a \acrfull{rpn} to first filter image regions containing objects from a set of search anchors. In a second stage, a \acrfull{roi} head extracts visual features from which it regresses enclosing bounding boxes as well as predicts a class label from a multi-set of classes for each object proposal in isolation. 
Classification is then learned using a cross-entropy loss across the confidence scores for each class, including a background class, where the class labels are found by matching the groundtruth bounding boxes to object proposal boxes using only the \acrfull{iou}. This formulation is vulnerable to imbalanced class distributions in the training data, therefore a focal loss \cite{Lin2020} down-weights well-classified examples in the cross-entropy computation to focus the training on hard examples.}
% , and an adaptive threshold for matching ground truth to proposal boxes \cite{Cai2018, Hosang2017, Liu2019c, Salscheider2020} was proposed to improve non-maximum suppression.
% These methods therefore require an additional \acrfull{nms} in order to prevent multiple detections of the same object, that use heuristics, like thresholds for overlapping, which in turn can be hard-coded~\cite{ren2015:faster-rcnn, Sun2020}, cascaded into a sequence of \acrshort{roi} heads \cite{Cai2018} trained with increasing \acrshort{iou} thresholds for proposal boxes used during the loss computation, or learned \cite{Hosang2017, Liu2019c, Salscheider2020}. 
% Recent object detection architectures attempt to incorporate scene context in the object detection process, either in the feature extractions stage, like DetectoRS~\cite{Qiao2020}, that integrates extra feedback connections into the backbone layers used for feature extraction, or on the proposal level by introducing context between object proposals.
% On Faster R-CNN architectures, graph neural networks extend the \acrshort{roi} heads and reformulate the object detection as a multi-set classification task~\cite{Yang2018, Hu2018}, that use a fully-connected graph in the \acrshort{roi} head for prediction refinement and duplicate removal, where nodes are proposals boxes and edge weights are derived from bounding box coordinates, classification score, and an encoded feature vector.

Anchor-less detectors such as keypoint-based or center-based methods~\cite{Tian2019, Zhou2019b, Law2020} embody an end-to-end approach that directly predicts objects as class-heatmaps. 
They use a logistic regression loss for training using multi-variate Gaussian distributions each centered at a groundtruth box. The classification stage has a global receptive field, which enables implicit modeling of context between detections. % CenterNet2~\cite{Zhou2021} bridges between the two concepts mentioned so far by relying on a keypoint-based method as region proposal network, and \acrshort{roi} head for refined regression and classification of each object detection. 
Transformer-based methods~\cite{Carion2020, Yang2021, dai2021dynamic} currently lead the \acrshort{coco}~\cite{Lin2014} object detection benchmark. These methods detect objects using cross-attention between the learned object queries and visual embedding keys, as well as self-attention between object queries to capture their interrelations in the scene context.
DETR~\cite{Carion2020} builds upon ResNet as a feature extractor and a transformer encoder to compute self-attention among all pixels of the feature map. However, the quadratic complexity w.r.t. the input resolution makes it infeasible for high-resolution feature maps, leading to poor performance on small objects. More efficient querying strategies, e.g., DeformableDETR \cite{Zhu2020}, was proposed that also support multiscale feature maps.\looseness=-1
% Research focus here is mainly on training practice, and reducing the computational complexity of the attention module w.r.t. the image resolution, for instance by focal self-attention~\cite{Yang2021} or focus attention on isolated dimensions, like scale level, height $\times$ width, and channels~\cite{dai2021dynamic}.	 

% While the latter methods incorporate context implicitly in the network design, we explicitly model context in order to analyse affordances using a scene graph formulation and knowledge graphs.

% \input{Graphics/tab_sota_results}

% MDETR \cite{Kamath} indicates that also natural language phrases can improve the grounding of object detectors and attribute classification.

%%%%%%%%%%%%%%%%%%%%%%%%%%%%%%%%%%%%%%%%%%%%%%%%%%%%%%%%%%%%%%%%%%%%%%%%%%%%%%%%%%%%%%%%%%%%%%%%%%%%%%
{\parskip=5pt
\noindent\textit{Knowledge-based embeddings}: In zero-shot classification~\cite{akata2015evaluation, Wang2018} and recognition~\cite{Bansal2018, Rahman2020, Gu2021, Zheng2021}, word embeddings commonly replace learnable class prototypes to transfer from training classes to unseen classes using inherit semantic relationships extracted from text corpora. 
Commonly used word embeddings are GloVe vectors~\cite{Wang2018,Bansal2018} and \textit{word2vec} embeddings~\cite{Rahman2020,Rahman2020a,Zheng2021,Li2021}, however embeddings learnt from image-text pairs using the CLIP \cite{Radford2021} achieve the best zero-shot performance so far \cite{Gu2021}. Rahman~\textit{et~al.}~\cite{Rahman2020} argue that a single word embedding per class is insufficient to model the visual-semantic relationships and propose to learn class representations of weighted word embeddings of synonyms and related terms \cite{Rahman2020}. Nevertheless, pure text embeddings perform consistently best for training classes~\cite{Bansal2018, Rahman2020, Gu2021, Zheng2021, Li2021, Gu2021} in object detection. 
The projection from visual to semantic space is done by a linear layer~\cite{Bansal2018, gupta2020multi, Li2021}, a single ~\cite{Rahman2020, Rahman2020a, Zheng2021} or two-layer MLP~\cite{Gu2021}, and learned with a max-margin losses \cite{Bansal2018, Rahman2020, NIPS2013_7cce53cf, gupta2020multi}, softplus-margin focal loss \cite{Li2021}, or cross-entropy loss \cite{Zheng2021,Gu2021}.
Zhang~\textit{et.~al.}~\cite{zhang2017learning} suggests to rather a mapping from semantic to visual space to benefit to alleviate the hubness problem in semantic spaces, however, our analysis of \acrshort{coco} class vectors suggested that the hubness phenomena does not have a noticeable impact on the \acrshort{coco} class embedding space. 
}

Another fundamental design choice in zero-shot object detection is the background class representation.
The majority of works rely on an explicit class prototype that is either learned~\cite{Zheng2021,Gu2021}, computed as the mean of all class vectors \cite{Rahman2020,Rahman2020a,gupta2020multi}, or represented by multiple background word embeddings \cite{Bansal2018}. However, Li~\textit{et~al.}~\cite{Li2021} noted that an explicit representation can cause confusions with unknown classes and therefore represented background by a distance threshold w.r.t. each class prototype. 

In this work, we analyze the usage of semantic relations between embeddings from the error distribution point of view and extend it to keypoint-based and transformer-based object detection architectures. We compare contrastive loss with margin-based loss functions and ablate multiple embedding sources to strengthen our analysis.

\section{Technical Approach}
\label{sec:technical}

\begin{figure*}
	\centering
	\begin{subfigure}[b]{\columnwidth}
		\centering
		\includegraphics[width=\textwidth]{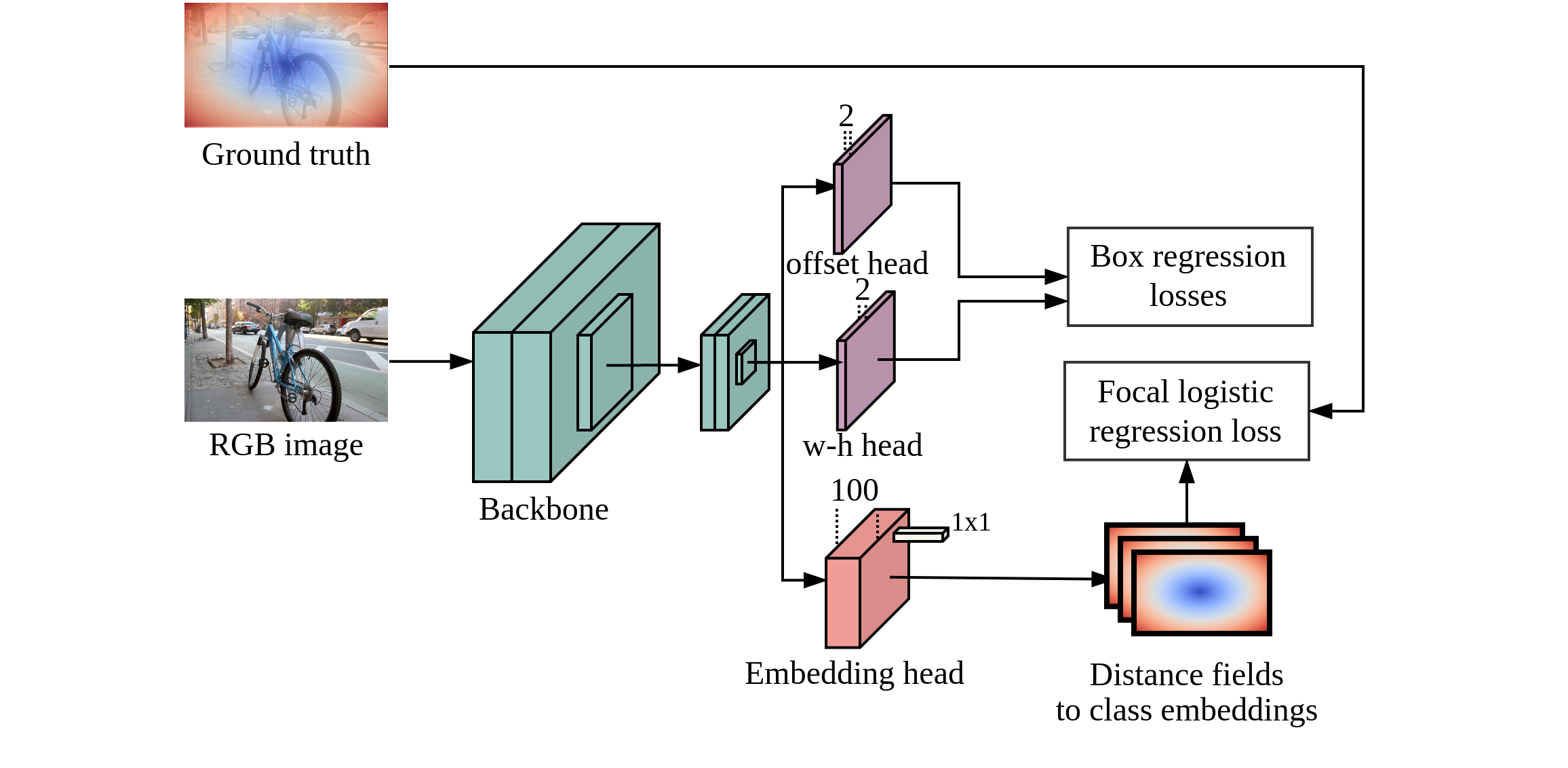}
		\caption{Knowledge graph embedded (KGE) CenterNet.}
		\label{fig:system_architecture_centernet}
	\end{subfigure}
	\begin{subfigure}[b]{\columnwidth}
		\centering
		\includegraphics[width=\textwidth]{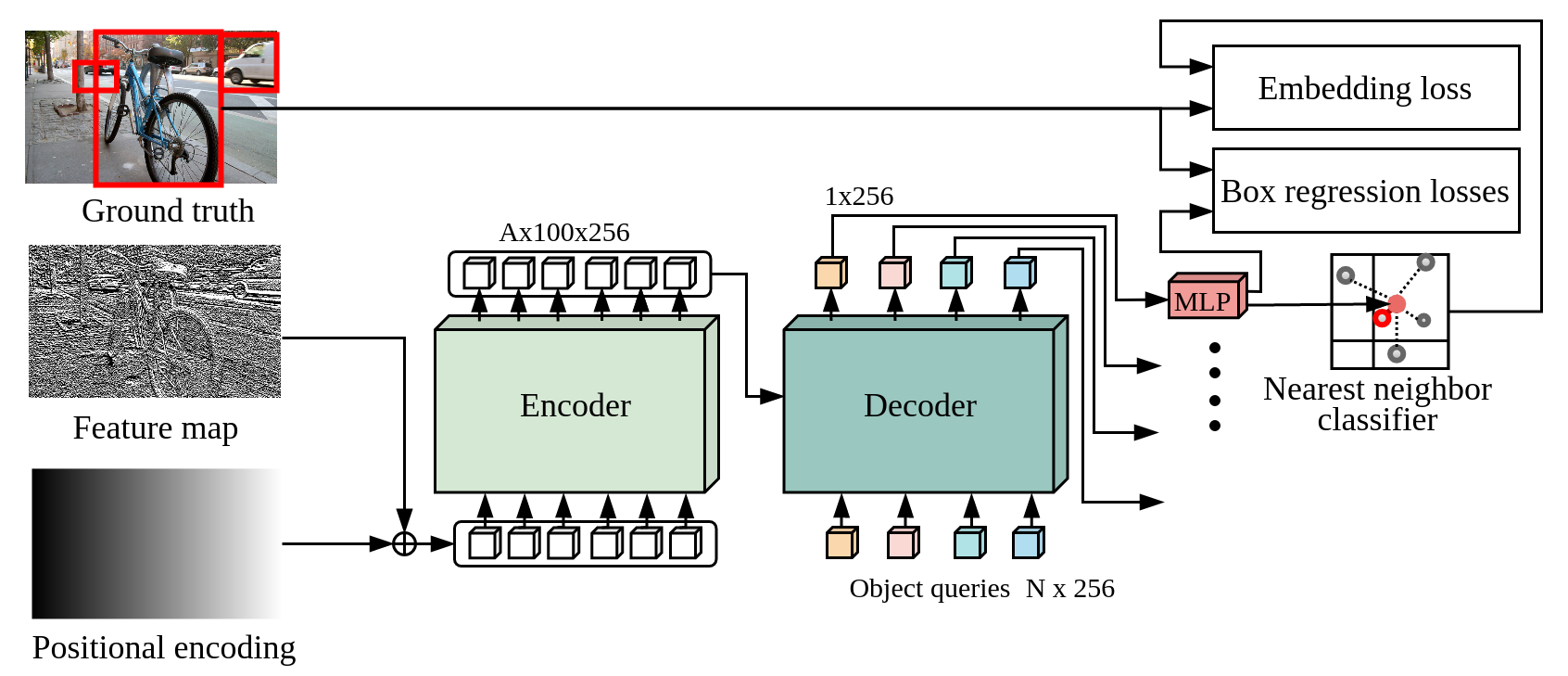}
		\caption{Knowledge graph embedded (KGE) DETR object detection head.}
		\label{fig:system_detr}
	\end{subfigure}%
	\label{fig:system_architecture}
	\vspace{-0.2cm}
	\caption{Knowledge graph embedded architecture using nearest neighbor classifiers for CenterNet architecture (a) and a transformer-head (b). 
		The CenterNet (a) encodes the image into a downscaled feature map from which it regresses a per-pixel class embedding vector, bounding box width and height, as well as centerpoint offset. 
		Object localization is performed by computing distance fields to each class prototype. For all the distances lower than a background threshold, we perform a local minima search in an 8-connected neighborhood to locate object center points.
		DETR (b) employs a transformer encoder and decoder with object queries as keys to detect objects. The decoded values are fed into an MLP that outputs bounding box coordinates as well as an embedding vector that is used for nearest neighbor classification given semantic class prototypes.
		}
	\vspace{-0.2cm}
\end{figure*}

Closed-set object detection architectures~\cite{ren2015:faster-rcnn,Zhou2019b,Carion2020,hurtado2020mopt} share a classification head $\mathbb{R}^D \mapsto \mathbb{R}^C$ design incorporating a linear output layer, that maps the feature vector into a vector of class scores of the form
\begin{equation}
    p(y = c | \mathbf{x}_i) = \textit{softmax}( \mathbf{W}^T \mathbf{x}_i)_c,
\end{equation}
where $\mathbf{W} \in \mathbb{R}^{D \times C}$ is the parameter matrix of the linear layer, which effectively is composed of C learnable class prototypes.
These parameters are learned with a one-hot loss formulation, incentivizing the resulting class prototypes to be pairwise orthogonal.

We replace learnable class prototypes with fixed object type embeddings that lend structure from external knowledge domains in the closed-set setting.
Essentially, this entails replacing the direct estimation of classification scores in object detection with a regression that maps image regions into a shared space of visual features and semantic embedding vectors, where classification is done by a nearest neighbor search.
% This novel formulation allows a structured representation of object types independent of the training data. 
In the remainder of this section, we present the possible class prototype choices in \secref{sec:object_representations}, distance metrics in \secref{sec:distance_metrics}, and describe the incorporation into two-stage detectors, keypoint-based method, and transformer-based architectures in \secref{sec:object_detector_integration}.
% The system is composed of a object detector and a graph neural network that allows to incorporate heterogeneous knowledge graphs into proposal box embeddings.
% The graph contains a classification edge type "isType", as well as edge types defining a spatial scene graph ("partOf", "nextTo").

\subsection{Object-Type Embeddings}
\label{sec:object_representations}

Instead of handcrafting the regression target vector or learning them from the training data, we incorporate object type embeddings from other knowledge modalities as follows: 
\begin{itemize}[topsep=0pt,noitemsep]
	\item \textit{GloVe} word embeddings~\cite{Pennington2014} build on the descriptive nature of human language learned from a web-based text corpus. Human language reflects a variety of dependencies between objects -including causal, compositional, as well as spatial interrelations- which is expected to reflect in the embeddings.
	\item \textit{ConceptNet} graph embeddings encode the conceptual knowledge between object categories entailed in the ConceptNet knowledge graphs~\cite{Speer2016}. The embeddings are computed by pointwise mutual information of the adjacency matrix and projected into a 100-D vector space by truncated \acrshort{svg}, as described by \textit{Speer et al.}~\cite{Speer2016}.
	\item \textit{COCO 2017} graph embeddings reflect the co-occurrence commonsense learned from the \acrshort{coco} training split. We build a heterogeneous graph from the spatial relations defined by Yatskar~\textit{et~al.}~\cite{Yatskar2016}: \{touches, above, besides, holds, on\} and transform the resulting graph into node embeddings for each object prototype with an R-GCN \cite{Woo2018}.
\end{itemize}

% The top-5 objects co-occurring with a query object type are represented in a homogeneous graph, for which a translational graph embedding is computed. 

\subsection{Distance Metrics for Classification}
\label{sec:distance_metrics}

We analyze nearest neighbor classification between visual embeddings of proposal boxes and object type representation vectors, either by the norm of the difference vector or angular distance. 
The $L_k$ norm of the difference vector between two embedding vectors is given by,
\begin{equation}
d_k(\overline{\mathbf{b}}_i, \overline{\mathbf{t}}_c) = \left(\sum_{d=1}^D \left(b_{i,d} - t_{c,d}\right)^k \right)^{1/k},
\end{equation}
where $\mathbf{b}_i \in \mathbb{R}^D$ denotes the feature vector of the \textit{i-th} proposal box and $\mathbf{t}_c \in \mathbb{R}^D$ the class prototype of the \textit{c-th} object class.
The maximum distance bound is given by the triangle inequality as $d_{k, max} \leq |\mathbf{t}_c|_k + |\mathbf{b}_i|_k$.
In order to have a bound independent of $D$, we project all embeddings inside the unit sphere as
\begin{equation}
    \overline{\mathbf{x}} = \frac{\mathbf{x}}{\max(1, |\mathbf{x}|_2)}.
\end{equation}

Since the relative contrast in higher dimensions is largest for low values of $k$~\cite{Aggarwal2001}, we analyze the Manhattan distance where $k=1$.
% Aggarwal~\textit{et.~al}~\cite{Aggarwal2001} found that the relative contrast in higher dimensions is largest for low values of $k$, and further propose fractional distance metrics where $k<1$ to increase the effectiveness of classification in high dimensions. For this work, we found a fractional norm with $k=\frac{2}{3}$ most accurate. 
We further investigate nearest neighbors in terms of the cosine angle between two embedding vectors. % , that for orthogonal vectors turns 0, while identical and opposite vectors result in 1 and -1 respectively. 
The distance is computed using the negated cosine similarity as
\begin{equation}
    d_{cos}(\mathbf{b}_i, \mathbf{t}_c) = 1 - \frac{\mathbf{b}_i^T \mathbf{t}_c}{|\mathbf{b}_i| |\mathbf{t}_c|}.
\end{equation}

We then interpret the negated distance as a similarity measure,  
\begin{equation}
    sim(\mathbf{b}_i, \mathbf{t}_c) = 1 - d(\mathbf{b}_i, \mathbf{t}_c) / 2
\end{equation} and define the embedding losses as a contrastive loss towards the class embedding vectors given by 
%\begin{equation}\resizebox{0.9\columnwidth}{!}{
%    $\mathcal{L}_{embd}(\mathbf{b}_i) = - sim(\mathbf{b}_i, \mathbf{t}_c) + \log \left( \sum^{C}_{i=0, i\neq c} \exp \left( sim(\mathbf{b}_i, \mathbf{t}_c) \right) \right).$}
%\end{equation}
\begin{equation}
\resizebox{.9\hsize}{!}{$\mathcal{L}_{embd}(\mathbf{b}_i, c_i) = - \log \left( 
    \frac{\exp \left( sim(\mathbf{b}_i, \mathbf{t}_{c_i}) / \tau \right)}{\sum^{C}_{c=1} \mathbb{1}_{[c\neq c_i] } \exp \left(sim(\mathbf{b}_i, \mathbf{t}_{k}) / \tau \right)}
    \right)$}
    \label{eq:contrastive_loss}
\end{equation}
where $c_i$ is the label of the groundtruth box to which the proposal is matched. $\tau$ denotes a fixed scaling factor of the similarity vector, commonly referred to as temperature. We follow \cite{Wang2020} and fix $\tau=0.07$.
% The dominator in \eqref{eq:contrastive_loss} can be seen to encourage representations to have large distances to class representation vectors in embedding space, while the nominator forces small distance to the true class representation vector.
In case of a feature vector $\mathbf{b}_i$ that is assigned to the background label, we set $c=0$ and $sim(\mathbf{b}_i, \mathbf{t}_0)=0$, as there is no class representation vector that the feature embedding should have close distance too. Please refer to the supplementary material for a comparison to a margin-based loss function.

\subsection{Generalization to Existing Object Detectors}
\label{sec:object_detector_integration}

In this section, we describe the embedding extension to three types of object detectors: two-stage detectors, keypoint-based methods, and transformer-based.

%%%%%%%%%%%%%%%%%%%%%%%%%%%%%%%%%%%%%%%%%%%%%%%%%%%%%%%%%%%%%%%%%%%%%%%%%%%%%%%%%%%%%%%%%%%%%%%%%%%%%%%%%%%%%%%%%%%%%%%%%%%%%%%%%%%%%%%%%%%%%%%%%%%%%%%%%%%%%%%%%%%%%%%%%%%%%%%%%%%%%%%
{\parskip=5pt
\noindent\textbf{Two-stage detectors}\label{sec:faster_rcnn_integration} such as R-CNN based methods~\cite{ren2015:faster-rcnn} pools latent features of each object proposals regions to fixed-size feature vector in the \acrshort{roi} head. These features are then processed in a box regressor and classifier to refine and classify the boxes for each proposal in isolation. The base method is trained with a standard Faster R-CNN loss, including  a cross-entropy loss for bounding box classification.
% During training, only proposals whose \acrshort{iou} with any of the ground truth boxes exceeds a threshold of 0.7 contribute to the loss computation as foreground objects.
For the Faster R-CNN \acrshort{kge} method, we replace the cross-entropy classification loss for an embedding loss function, such that the linear layer learns a regression towards fixed class prototype vectors rather than mutually optimizing both. We therefore task the linear output layer to map from the \acrshort{roi} feature dimensionality to the  embedding dimensionality $D$. We replaced the ReLU activation at the box head output layer with a hyperbolic tangent activation function (tanh) to avoid zero-capping the \acrshort{roi} feature vectors before the final linear layer.}

% The embeddings are learned with a supervised contrastive loss to all object type representation vectors as
% \begin{equation}
%   \mathcal{L}_{i,j} = -log \frac{exp\left(sim(\mathbf{z}_i, \mathbf{z}_j) / \tau\right)}{\sum_{k=1, k\neq i}^{C} exp\left(sim(\mathbf{z}_i, \mathbf{z}_k) / \tau\right)},
% \end{equation}
% where i refers to a proposal box, and $sim$ is a similarity metric like euclidean distance or cosine similarity in embedding space.

% We employ the fine-tuning strategy proposed in~\cite{Sun2021}, and, starting from a pre-trained model, freeze the feature extractor backbone, and re-training the \acrshort{rpn} and \acrshort{roi} modules, as shown in \figref{fig:system_fasterrcnn}

%%%%%%%%%%%%%%%%%%%%%%%%%%%%%%%%%%%%%%%%%%%%%%%%%%%%%%%%%%%%%%%%%%%%%%%%%%%%%%%%%%%%%%%%%%%%%%%%%%%%%%%%%%%%%%%%%%%%%%%%%%%%%%%%%%%%%%%%%%%%%%%%%%%%%%%%%%%%%%%%%%%%%%%%%%%%%%%%%%%%%%%
{\parskip=5pt
\noindent\textbf{Keypoint-based detectors} \label{sec:centernet_integration} formulates object detections as a keypoint estimation problem. CenterNet~\cite{Zhou2019b} uses these keypoints as the center of the bounding boxes. The output stage has a convolutional kernel with filter depth equal to the number of object classes, as well as additional output heads for center point regression, bounding box width, and height, etc. Max-pooling on class-likelihood maps yields a unique anchor for each object under the assumption that no two bounding boxes share a center point. The class-likelihood maps are trained with a focal logistic regression loss, where each groundtruth bounding box is represented by a bivariate normal distribution $Y_{xyc}$ with mean at the groundtruth center point and class-specific variances.}
% This formulation avoids the non-maximum suppression reduces to the problem of finding local maxima in a 8-connected neighborhood, which makes the anchor search differentiable. 
% Another benefit of the fully convolutional formulation is that each pixel in the keypoint map has a global receptive field, thus the network could inherently learn to reason from scene context of the entire picture.

For the \acrshort{kge} formulation, we use the filter depth of the embedding output stage according to the number of embedding dimensions, thus learning an embedding for each pixel in the output map as shown in \figref{fig:system_architecture_centernet}. Center point search then requires to compute a distance field of each pixel embedding to the class representation vectors. We formulate the embedding based focal loss as
\begin{equation}
\resizebox{.9\hsize}{!}{$\mathcal{L}_{focal} = 
\begin{cases}
(1-sim(\mathbf{b}_{x,y}, \mathbf{t}_c))^{\alpha} \log \left(sim(\mathbf{b}_{x,y}, \mathbf{t}_c)\right ))Y_{xyc}^{\beta} & \text{ if } Y_{xyc} =< 1 \\ 
sim(\mathbf{b}_{x,y}, \mathbf{t}_c)^{\alpha} \log \left(  1 - sim(\mathbf{b}_{x,y}, \mathbf{t}_c)\right )) & \text{ otherwise }
\end{cases}$}
\end{equation}
where $sim(\mathbf{b}_i, \mathbf{t}_c)$ is a similarity function, $\alpha$ and $\beta$ are hyperparameters of the focal loss, and $Y_{xyc}$ is a heatmap where each groundtruth bounding box of class $c$ is represented by a bivariate normal distribution $Y_{xyc}$ at its center point coordinates $x$ and $y$, scaled such that $Y_{xyc}=1$.
% For the calculations of the losses, we derive classifications scores for each object type by convolving the target representation vector of the heatmap, and compute a distance function for each pixel, as follows
%\[
%    score(c, x, y) = dist(\mathbf{t}_c, \mathbf{F}_{x, y}) \
%\]
%for class $c$ and pixel location $(n, m)$.

%%%%%%%%%%%%%%%%%%%%%%%%%%%%%%%%%%%%%%%%%%%%%%%%%%%%%%%%%%%%%%%%%%%%%%%%%%%%%%%%%%%%%%%%%%%%%%%%%%%%%%%%%%%%%%%%%%%%%%%%%%%%%%%%%%%%%%%%%%%%%%%%%%%%%%%%%%%%%%%%%%%%%%%%%%%%%%%%%%%%%%%
{\parskip=5pt
\noindent\textbf{Transformer-based methods}~\cite{Carion2020,Zhu2020} adopt an encoder-decoder architecture to map a ResNet-encoded image to features of a set of object queries.
The DETR encoder computes self-attention of pixels in the ResNet output feature map concatenated with a positional embedding. 
% The input of each attention layer is concatenated with a learnt positional encoding and transformed into N output feature vectors by the decoder.
The decoder consists of cross-attention modules that extract features as values, whereby the query elements are of N object queries and the key elements are of the output feature maps from the encoder. These are followed by self-attention modules among object queries.
Each feature vector is then independently decoded into prediction box coordinates and class labels with a 3-layer perceptron with ReLU activation~\cite{Carion2020}, as depicted in \figref{fig:system_detr}. 
The network is trained with a unique matching between proposal boxes and groundtruth boxes based on a classification cost with the negative log-likelihood of class labels, a generalized \acrshort{iou} loss as well as L1 regression loss on the bounding box coordinates.}
% DETR uses a ReseNet-50 backbone to encode the input image and processes the resulting feature map by a transformer with multi-head self- as well as encoder-decoder attention and finally collecting the multi-head attention output. 

% The quadratic complexity with the inpute size makes DETR infeasible for high resolution feature maps that e.g. facilitate to detect small objects.
% Furthermore, DETR requires many training epochs a factor of 5 more epochs to converge compared to Faster R-CNN or CenterNet architectures, assumably due to the attention maps that at initalization are almost of average attention on the whole feature map while towards the end the attention maps become comparable sparse.

% \usepackage{multirow}
% \usepackage{graphicx}
\begin{table*}
\footnotesize
\centering
\begin{tabular}{llll|ccc|p{0.4cm}p{0.5cm}p{0.5cm}p{0.5cm}p{0.5cm}p{0.4cm}}
\toprule
                                       &                   &                          &                 & \multicolumn{3}{c|}{{COCO 2017 val}} & \multicolumn{6}{c}{{COCO 2017 test-dev}}                                                               \\
\cmidrule{5-13}{Model}          & {Backbone} & {Prediction Head} & {Params} & AP & ${\text{AP}_w}$     & ${\text{AP}_{cat,w}}$  & {AP}   & ${\text{AP}_{50}}$ & ${AP_{75}}$ & ${\text{AP}_S}$ & ${\text{AP}_M}$ & ${\text{AP}_L}$ \\ \midrule
Faster R-CNN~\cite{ren2015:faster-rcnn} & R50-FPN           & Fast R-CNN               & 42 M            & 40.2 & 40.9            & 41.9                      & 40.2          & 61.0               & 43.8               & 24.2            & 43.5            & 52.0            \\
Faster R-CNN  \acrshort{kge}            & R50-FPN           & ConceptNet-Cossim        & 42 M            & 39.3 & 41.6            & 43.5                      & 39.4          & 59.7               & 43.4               & 24.1            & 41.3            & 49.8            \\
\midrule
CenterNet~\cite{Zhou2019b}             & DLA               & $3\times3$ Conv + FFN    & 19 M            & 42.5 & 43.9            & 45.1                      & 41.6          & 60.3               & 45.1               & 21.5            & 43.9            & 56.0            \\
CenterNet \acrshort{kge}               & DLA               & GloVe-Cossim             & 19 M            & \textbf{44.2} & \textbf{45.5}            & \textbf{47.3}                      & \textbf{44.6}          & 63.6               & \textbf{49.2}      & \textbf{29.4}   & \textbf{47.0}            & 53.5            \\ \midrule % ---------------------------------------------------------------------
%Faster RCNN  \acrshort{kge}            & R50-FPN           & %ConceptNet-Cossim        & 42 M            & 39.7            & 42.1                      & 37.3          & 57.8               & 40.7               & 21.2            & 39.8 & 46.5 \\
%CenterNet \acrshort{kge}               & DLA               & %GloVe-Cossim             & 19 M            & 43.8            & 44.8                      & 41.4          & 59.3               & \textbf{45.6}      & \textbf{24.9}   & 45.0            & 50.4            \\
%DETR \acrshort{kge}                    & R50-FPN           & ConceptNet-Cossim        & 41 M            & 41.5            & \textbf{42.6}             & 41.6          & 61.3               & 44.5               & 20.2            & 44.4            & 58.1 \\ \midrule % ---------------------------------------------------------------------
DETR~\cite{Carion2020}                 & ResNet-50         & 3-layer FFN              & 41 M            & 42.0 & 40.3             & 41.2                       & 42.0 & 62.4      & 44.2               & 20.5            & 45.8   & \textbf{61.1}   \\
DETR \acrshort{kge}                    & R50-FPN           & ConceptNet-Cossim        & 41 M            & 43.2 & 42.9            & 43.6             & 43.3          & \textbf{65.3}               & 45.8               & 22.9            & 45.9            & 59.8            
\\ \bottomrule
\end{tabular}
	\vspace{-0.2cm}
\caption{COCO 2017 \textit{test-dev} and \textit{val} results.
The top section shows baseline results with learnable class prototypes and a one-hot loss, where \textit{test-dev} results were taken from the referenced literature and \textit{val} results were reproduced using their \textit{detectron2} implementations.
The bottom section presents results using test time augmentations for the proposed regression heads that vary in the type of class prototypes representation. For each meta architecture, we chose the best performing prediction head on the validation set. The COCO 2017 \textit{val} columns show the average precision of predicting the correct class or category, which we could only evaluate on the validation set.}
\label{tab:results_val_coco}
\vspace{-0.2cm}
\end{table*}

We propose to replace the class label prediction with a feature embedding regression, and replace the classification cost function in the Hungarian matcher by the negative logarithm of the similarity measure, as described in \secref{sec:distance_metrics}. We further replace the ReLU activation function of the second-last linear layer with a tanh activation function.

%the transformer encoder-decoder architecture computes self-attention between all image regions, and therefore takes the global image context into account when predicting object proposals.

%The generalization to an embedding vector is straight-forward to the Two-stage method, and we additionally estimate a background target

\section{Experimental Evaluation}
\label{sec:experiments}
We evaluate the nearest neighbor classification heads for 2D object detection on the \acrshort{coco} and Cityscapes benchmarks. We use the mean average precision ($AP$) metric averaged over IoU $\in [0.5 : 0.05 : 0.95]$ (COCO’s
    standard metric) as the primary evaluation criteria. We also report the  $AP_w$ where we weight each class by the number of groundtruth instances, the average precision across scales $AP_{S, M, L}$ and the categories $AP_{cat}$ for completeness.
    The latter is a novel metric first introduced in this paper, which is based on categorizing the 80 \acrshort{coco} \textit{thing} classes into a total of 19 base categories by finding common parent nodes based on WUP similarity in the Google knowledge graph \cite{google_knowledge_graph}. A true positive classification label hereby defines the correct category rather than class. Please refer to the supplementary material for an overview of object categories.

    We chose the object detection task over image classification as it additionally requires predicting whether a proposed image region contains a foreground object.
    Ablation studies are performed on the validation set, and a architecture-level comparison is reported on the \acrshort{coco} \textit{test-dev} and Cityscapes \textit{val} datasets. 
    For object detection algorithms, we compare Faster R-CNN \cite{ren2015:faster-rcnn} as representative for anchor-based methods, CenterNet \cite{Zhou2019b} as keypoint-based object detector, and DETR~\cite{Carion2020} as transformer-based representative. 
    
% \usepackage{graphicx}
\begin{table}
\centering
\footnotesize
\begin{tabular}{l|lllll}
\toprule
  & \multicolumn{5}{c}{{Cityscapes val}}   \\
\cmidrule{2-6}
{Model}                 & {AP} & ${\text{AP}_{50}}$ & ${\text{AP}_{75}}$ &${\text{AP}_{w}}$ &${\text{AP}_{cat}}$\\ \midrule
Faster R-CNN~\cite{ren2015:faster-rcnn} & 41.4 & 62.5 & 43.4 & 48.5 & 49.3 \\ 
% CenterNet~\cite{Cai2018} & - & - & - & -\\
% DETR~\cite{Carion2020}   & - & - & - & -\\
\midrule
Faster R-CNN \acrshort{kge} & 34.5 & 58.3	& 34.4 & 41.4 & 41.8 \\
CenterNet \acrshort{kge}    & \textbf{42.2} & \textbf{63.4} & \textbf{44.4} & \textbf{51.7} & \textbf{52.2} \\ 
DETR \acrshort{kge}         & 39.7 & 61.1 & 42.5 & 47.2 & 47.9 \\ \bottomrule
\end{tabular}
	\vspace{-0.2cm}
\caption{Validation set results for the Cityscapes dataset.
The top section shows baseline results in detectron2 with standard one-hot classification heads. The bottom section shows results for the proposed regression heads for the same configurations as in \tabref{tab:results_val_coco}.}
\label{tab:cityscapes_test_results}
	\vspace{-0.3cm}
\end{table}
    
    \subsection{Datasets}
    
    % We evaluate our proposed nearest-neighbor classification approach on two datasets, the \acrshort{coco} and cityscapes dataset.
    
    % Object type representations from various sources and modalities are compared:  hosen as knowledge graph embeddings from ConceptNet, GloVe word embeddings, and co-occurrence embeddings from the COCO dataset train split.
    
    \noindent\textit{COCO} is currently the most widely used object detection dataset and benchmark. The data depicts complex everyday scenes containing common objects in their natural context. Objects are labeled using per-instance segmentation to aid in precise object localization \cite{Lin2014}.
    We use annotations for 80 "thing" objects types in the 2017 train/val split, with a total of 886,284 labeled instances in 122,266 images.
    The scenes range from dining table close-ups to complex traffic scenes.

    \noindent\textit{Cityscapes} is a large-scale database that focuses on semantic understanding of urban street scenes. It provides object instance annotations for 8 classes of around 5000 fine annotated images captured in 50 cities during various months, daytime, and good weather conditions \cite{Cordts2016}. 
    The scenes are extremely cluttered with many dynamic objects such as pedestrians and cyclists that are often grouped near one another or partially occluded.
    
    \subsection{Training Protocol}
    
    % For the following experiments, the detectron2 \cite{Wu2019a} library was used.

    % The models were trained with the same training schedules as the baseline methods, whereby the architecture replaced with the embedding modules described in \ref{sec:object_detector_integration} was used.
    % We were using network parameters pre-trained on the respective datasets using the standard \acrshort{roi} heads, but replace the \acrshort{roi} head with the link prediction modules presented in \ref{sec:link_predictor}.
    % The models were fine-tuned on 5 epochs on the respective datasets using stochastic gradient descent and a three-step learning schedule.
    % The learning rate schedule disabled gradients for the backbone, \acrshort{fpn}, and \acrshort{rpn} for the two epochs, in order for the link predictor to converge towards a graph embedding, afterwards all network parameters were updated by the training.
    % After two and four epochs respectively, the learning rate for all modules is reduced by one-tenth.
    % \todo{Plot class error distribution}
    
    We use the PyTorch~\cite{NEURIPS2019_9015} framework for implementing all architectures, and we train our models on a system with an Intel Xenon@2.20GHz processor and NVIDIA TITAN RTX GPUs. 
    For comparability, we use a ResNet-50 backbone with weights pre-trained as an image classifier on the ImageNet dataset for all experiments, and the same training configurations as reported in \cite{ren2015:faster-rcnn} for Faster R-CNN \acrshort{kge}, in \cite{Zhou2019b} for CenterNet \acrshort{kge}, and as in \cite{Carion2020} for DETR \acrshort{kge} configurations.
    We rescale the images to $640\times480$ pixels and use random flipping as well as cropping for augmentation.

    \subsection{Quantitative Results}
    
    % Table  shows the results on the \acrshort{coco} test set.
    % The results on the cityscapes dataset are reported in table \ref{tab:cityscapes_test_results}.
    
    On the \acrshort{coco} test set in \tabref{tab:results_val_coco}, we compare our results against baselines from the original methods presented in \secref{sec:object_detector_integration}.
    % Additionally, we provide test set accuracies for Cascade R-CNN \cite{Cai2018} as a straight-forward upgrade of the Faster R-CNN (KGE) detectors.
    % ; and DetCo \cite{Xie2021}, a method using self-supervised pre-training on global as well as local patch augmentations and a contrastive loss function with the aim of structuring the feature space rather than the classification stage. 
    We expected the baselines methods to outperform our proposed \acrfull{kge} prediction heads since they favor one-hot encoded class prototypes which are pairwise orthogonal and therefore maximize inter-class distances.
    Nevertheless, \tabref{tab:results_val_coco} shows that the knowledge-based class prototypes in the final classification layer can compete with their baseline configurations in terms of the $AP$ metrics. On the \acrshort{coco} \textit{test-dev} benchmark, the CenterNet \acrshort{kge} outperforms its one-hot encoded counterparts by $+1.7\%$ , while the DETR \acrshort{kge} variant achieves an overall $AP$ of $43.2$. The Faster R-CNN \acrshort{kge} method does not outperform its baseline performance in average precision, however,  the comparison of the $AP_w$ metric shows that this could be dependent on the test set's class-distribution. The precision gain is largest for $AP_{75}$ over all the methods, where the \acrshort{kge} methods benefit from high accuracy bounding boxes.
    
    We attribute this competitive performance  of the \acrshort{kge} methods to the small temperature value of $\tau=0.07$, which was used as a hyperparameter for the contrastive loss function. As Kornblith~\textit{et~al.}~\cite{Kornblith2020} noted, the temperature controls a tradeoff between the generalizability of penultimate-layer features and the accuracy on the target dataset. We performed a grid search over temperature parameters and found that accuracy drops considerably with larger temperature values, as shown in the supplementary material. Another interesting observation are the large $AP_{category,w}$ scores, where the \acrshort{kge} methods consistently outperform their baselines by greater than $1.5\%$. These results demonstrate that misclassifications are more often within the same category, in contrast to baseline methods that appear to confuse classes across categories.
    
    The results on the Cityscapes \textit{val} set are shown in \tabref{tab:cityscapes_test_results}. The CenterNet \acrshort{kge} variant achieves the best precision with $AP=42.2\%$. We should note, that for this dataset with only few classes that are highly related, the Faster R-CNN \acrshort{kge} methods performs worse than its baseline, mainly due to few accurately predicted bounding boxes as can be seen by the low $AP_{75}=34.4$. 
    % We interpret this metric an indicator of semantic context in the misclassifications of a detector.
    % We argue that a class confusion within these categories is less fatal than a confusion with a class of another category, since members of the same category are expected to share a significant part of their behaviors and appearances that are important when interacting with them.

    % Please note that $\Delta$ can be negative even if the categories are chosen to have cardinality $\geq 1$, because $AP$ is computed as an unweighted average of all classes, whereas $AP_{category}$ is weighted by the class occurrences.
   
    % However, we note that the $AP_{category}$ for CenterNet is lower than its CenterNet \acrshort{kge} counterpart, in fact the CenterNet baseline has  a considerably high $\Delta$ compared to other one-hot encoded baselines.

    % The improvements are consistent over all object sizes, therefore the knowledge-embedding seem not to provide additional information to recognize object within small or partially visible objects.

    \subsection{Ablation Study}
    
    In this section, we compare three class prototype representations described in \secref{sec:object_representations} to analyze which aspects of semantic context are most important for the object detection task and evaluate distance metrics used for classification. 
    
% \usepackage{graphicx}
\begin{table}
\centering
\footnotesize
\begin{tabular}{llll|p{0.6cm}p{0.6cm}p{0.6cm}}
\toprule
{Model} & \multicolumn{3}{l|}{{Class Prototypes}}                                                                                                   & \multicolumn{3}{c}{{COCO 2017 val}}                                                                        \\
                                & \rotatebox[origin=l]{90}{GloVe} & \rotatebox[origin=l]{90}{ConceptNet} & \rotatebox[origin=l]{90}{COCO 2017} & \multicolumn{1}{c}{{AP}} & \multicolumn{1}{c}{${AP_{50}}$} & \multicolumn{1}{c}{${AP_{75}}$} \\
                                \midrule
% Faster R-CNN~\cite{ren2015:faster-rcnn} &   &  &    & 40.22                  & 61.02                                  & \textbf{43.81}                                  \\
Faster R-CNN KGE                    & x                                                      &                                                             &                                                            & 38.24                  & 57.49                                  & 41.98                                  \\
                                &                                                        & x                                                           &                                                            & 40.39         & \textbf{61.98}                         & 41.58                         \\
                                &                                                        &                                                             & x                                                          & 31.40                  & 47.33                                  & 34.84                                  \\ \midrule
% CenterNet~\cite{Zhou2019b}                    &   &  &    & \textbf{42.51}                  & \textbf{59.95}                                  & 46.23                                  \\
CenterNet KGE                    & x  &   &  & 41.37         & 59.23                         & \textbf{45.24}                         \\
                                 &  & x   &  & 41.04                 & 58.76                                  & 44.97                                  \\
                                &                                                        &                                                             & x                                                          & 41.07                  & 58.94                                  & 45.09                                  \\ \midrule
% DETR~\cite{Carion2020}                   &   &  &    & \textbf{42.00}                  & \textbf{62.32}                                  & \textbf{44.29}                                  \\
DETR   \acrshort{kge}           & x  &  & & 40.29                & 59.30                                & 42.73                                  \\
                                &  & x  & & \textbf{41.40}         & 61.20                         & 43.92                         \\
                                &  &  & x & 37.36                  & 54.85                                  & 39.53                                  \\
\bottomrule
\end{tabular}
	\vspace{-0.2cm}
\caption{Ablation study on \acrshort{coco} validation set results over knowledge embeddings in 100 dimensions using a cossim distance metric and a contrastive loss (\eqref{eq:contrastive_loss}).
The class prototypes are described in \secref{sec:object_representations}}
\label{tab:results_ablation_embeddings}
\end{table}
    
    {\parskip=5pt
    \noindent\textbf{Analysis of Object Type Representations}: \tabref{tab:results_ablation_embeddings} shows the average precision for the object detection architectures presented in \secref{sec:object_detector_integration} when using various class prototype representations on the \acrshort{coco} \textit{val} set. The ConceptNet embedding gives the best $AP$ over the Faster R-CNN \acrshort{kge} and DETR \acrshort{kge} configurations, however the results for using a GloVe embedding performs best for the CenterNet architecture. The COCO embeddings consistently perform the worst for all \acrshort{kge} configurations. This is exceptional, since the COCO embeddings summarize the co-occurrence statistics without domain shift, while the ConceptNet and GloVe embeddings introduce external knowledge from textual and conceptual sources. These results indicate that the semantic contexts' role is dominated by categorical knowledge about the object classes. Context from co-occurrence statistics appears to generate insufficient class separation for object types that are frequently overlapping, etc., resulting in a comparably large gap for Faster R-CNN when comparing COCO 2017 to the remaining embedding types. % We assume that this effect is especially severe for Faster R-CNN since this architecture uses a groundtruth assignment based on overlap with of bounding boxes for loss computation.
    }

% \usepackage{graphicx}
\begin{table}
\centering
\footnotesize
\begin{tabular}{lp{0.7cm}p{0.7cm}|rrr}
\toprule
{Model} & \multicolumn{2}{c|}{{Distance Metric}}                                                                                                                                                                                                                   & \multicolumn{3}{c}{{COCO 2017 val}}                                                                        \\
                                & \rotatebox[origin=l]{90}{\thead{Cosine\\similarity}} & \rotatebox[origin=l]{90}{\thead{Manhattan\\distance}} & \multicolumn{1}{c}{{AP}} & \multicolumn{1}{c}{${AP_{50}}$} & \multicolumn{1}{c}{${AP_{75}}$} \\ \midrule
Faster R-CNN   \acrshort{kge}              & x                                                                                                                          &                                                                                                                           & 40.39         & \textbf{61.98}                         & 41.58                         \\
                                &                                                                                                                            & x                                                                                                                         & 35.99                  & 56.88                                  & 38.87                                  \\ \midrule
CenterNet   \acrshort{kge}                    & x                                                                                                                          &                                                                                                                           & 41.37        & 59.23                       & \textbf{45.24}                        \\
                                &                                                                                                                            & x                                                                                                                         & 16.42  & 22.97                   & 18.57                  \\ \midrule
DETR   \acrshort{kge}                         & x                                                                                                                          &                                                                                                                           & \textbf{41.40}         & 61.20                         & 43.92                         \\
                                &                                                                                                                            & x                                                                                                                         & 40.37                  & 60.16                                  & 42.70                                   \\ \bottomrule
\end{tabular}
	\vspace{-0.2cm}
\caption{Ablation study on \acrshort{coco} validation set results over distance metrics for configurations using ConceptNet embeddings in 100 dimensions as class prototypes.}
\label{tab:results_ablation_distance}
\vspace{-0.4cm}
\end{table}

    {\parskip=5pt
    \noindent\textbf{Analysis of Distance Metric}: In \tabref{tab:results_ablation_distance}, we further compare validation set results for Manhattan and cosine distances in the nearest neighbor classification. The latter achieves consistently higher average precision values over all the architectures. The gap is largest for the CenterNet architecture, where the cosine similarity variant achieves $+24.7$ higher $AP$ and lowest for DETR with $+1.02 AP$. This might reflect the sensitivity to outliers for each method and the separation of the embedding space. In high dimensions, the euclidean distance between two points is inherently large, since the vector becomes increasingly sparser. According to Aggarwal~\textit{et~al.}\cite{Aggarwal2001}, this results in decreasing contrast by the $L_k$-norm distance between embedding vectors with increasing dimensionality, which hurts classification performance. The cosine similarity on the other hand reports the same similarity between each point on a line from the origin to a target point, in principle reserving the scaling dimension to account for different appearances of an object class. Further, the effect of outlier values in a single dimension only enters the distance metric at most linearly.}
    
    % Hence, the euclidean distance can enforce stronger partitioning of the classification space when used in the loss function, especially in high dimensions. This however results in large "white spaces" in the embedding space, that react noisy to the nearest neighbor search.
    % Cosine similarity on the other hand is much less sensitive to the dimensionality of the feature space and more moderate in increasing inter-class distance. However, the gradient flow through a loss using cosine similarity is considerable less steep, than when using euclidean distance, which could result in slow convergence or increased the number of local minima one the cost function plane.
    
    % \input{tables/results_ablation}

    \subsection{Misclassification Analysis}
    
    We analyze the confusion matrix of each detector on the \acrshort{coco} validation split.
    The confusion matrix $\mathbf{E} \in \mathrm{R}^{C \times C}$ contains a row for each object class, and each entry represents the counts of predicted labels for this class groundtruth instances.
    For each groundtruth box, we select the predicted bounding box with the highest confidence score of all predicted boxes with an \acrshort{iou} $\geq 0.8$, i.e., each groundtruth box is assigned to at most one prediction.
    
    For comparing error distributions of different object detectors, we compute the Jensen-Shannon distance of corresponding rows in the confusion matrices as
    \begin{equation}
        JS(\mathbf{p}, \mathbf{q}) = \sqrt{\frac{D(\mathbf{p} \parallel \mathbf{m}) + D(\mathbf{q} \parallel \mathbf{m})}{2}},
    \end{equation}
    where $\mathbf{p}$, $\mathbf{q}$ are false positive distributions, $\mathbf{m}$ is the point-wise mean of two vectors as $m_i = \frac{p_i + q_i}{2}$, and $D$ is the Kullback-Leibler divergence.
    
    % as
    % \begin{equation}
    %     D(\mathbf{p} \parallel \mathbf{m}) = \sum_i p_i \log \left( \frac{p_i}{m_i}\right).
    % \end{equation}
    
\begin{figure}
        \centering
        \includegraphics[width=0.8\linewidth]{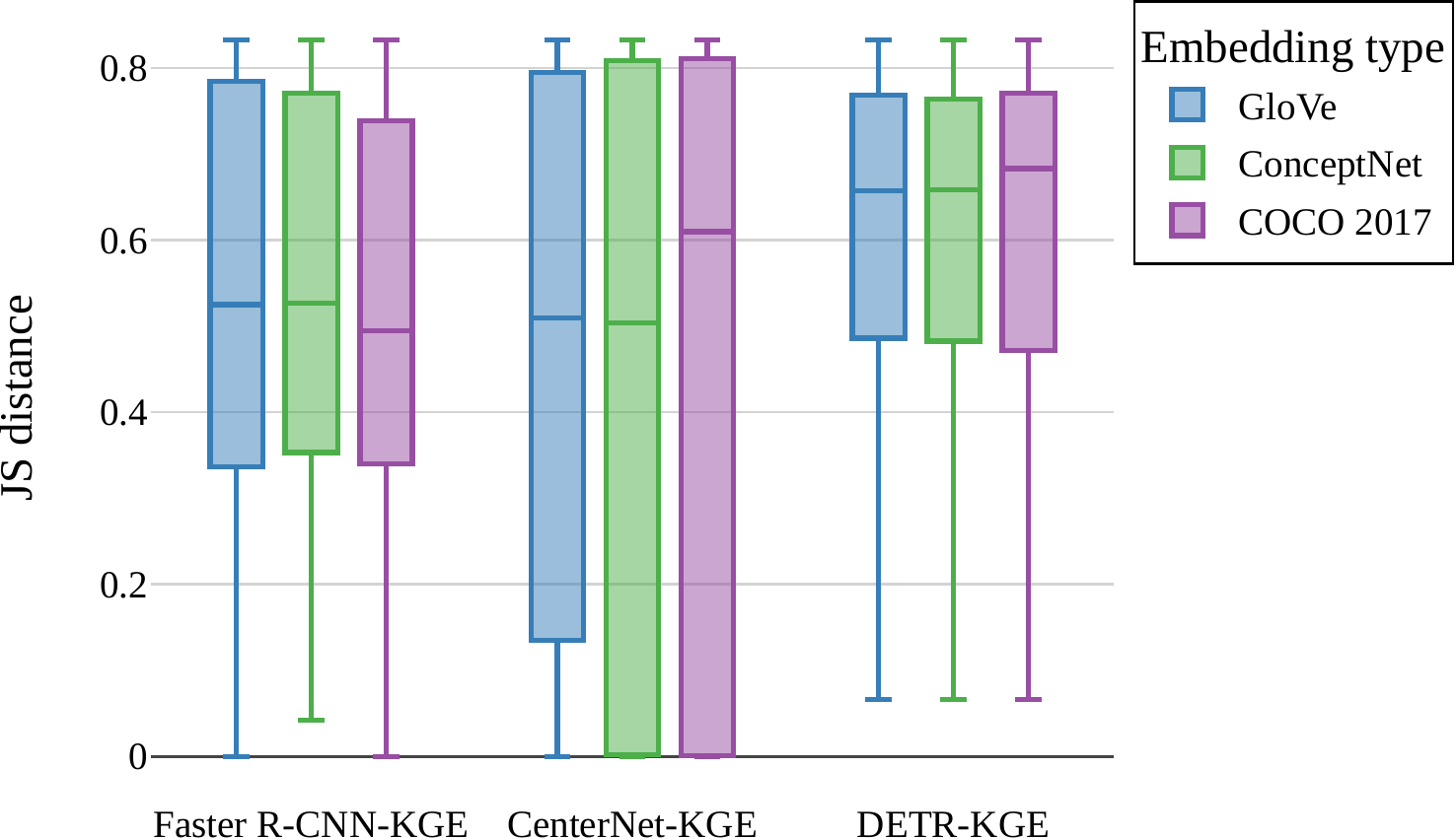}
        \caption{Jensen-Shannon distance between classification error distributions of detector type in baseline configuration compared to its variants using the knowledge-embedded class-prototypes described in \secref{sec:object_detector_integration}.}
        \label{fig:js_distances_coco}
        \vspace{-0.2cm}
    \end{figure}

\begin{figure}
        \centering
        \includegraphics[width=0.8\columnwidth]{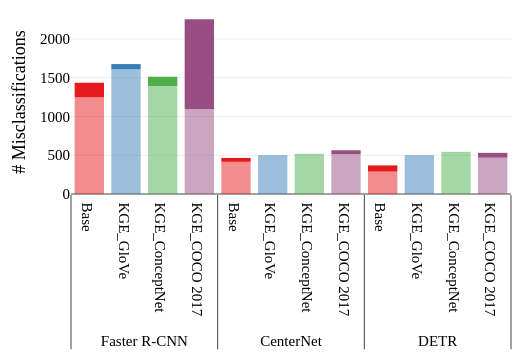}
        \caption{Misclassifications on the \acrshort{coco} validation split for each detector configuration. 
        % A groundtruth bounding box is misclassified, if the predicted bounding box with the highest IoU indicates any other than the ground truth label. 
        Light colors represent if the misclassified label has a common node with the groundtruth class in the Google knowledge graph \cite{google_knowledge_graph} (intra-category confusion), solid colors indicate that there is no common parent node (inter-category confusion).}
        \label{fig:supercategory_confusion}
        \vspace{-0.2cm}
    \end{figure}

    A low JS distance implies that the classification stages of the two detectors under comparison produce similar error distributions and vice versa. \figref{fig:js_distances_coco} shows the JS distances for each object detection algorithm between the baseline method and all embedding-based prediction heads. We note that the error distributions vary noticeably from the baselines methods, however all the embedding-based prediction heads appear to differ by similar magnitudes. This behavior indicates that there is a conceptual difference in the prediction characteristic of the learned (baseline) and knowledge-based class prototypes. The effect is largest for the CenterNet architecture, presumably since the embedding-based formulation affects the classification as well as localization head. To further investigate the origin of this dissimilar behavior, we quantify the inter-category confusions of different object detector architectures in \figref{fig:supercategory_confusion}. 
    
    {\parskip=5pt
    \noindent\textbf{Inter-Category Confusion}: \figref{fig:supercategory_confusion} shows that all knowledge-embedded class prototypes, except for the \textit{COCO 2017} class prototypes, produce lower inter-category confusions compared to their one-hot encoded configurations.
    This signifies that feature embeddings derived from spatial relations between objects provide insufficient inter-class distances when used as class prototypes. This impairment is especially noticeable for the Faster R-CNN architecture, where class confusion of overlapping objects can also occur during the loss computation.
    The ConceptNet embeddings result in the lowest counts of inter-category confusion, since its structure is derived from a conceptual knowledge graph similar to the one used for categorization, and its inter-category confusion counts are consistently lower than the base configuration with learnable class prototypes.}

    \subsection{Qualitative Results}

\begin{figure*}
    \centering
    \footnotesize
    \setlength{\tabcolsep}{0.05cm}% for the horiz padding
    {\renewcommand{\arraystretch}{0.15}% for the vertical padding
    \newcolumntype{M}[1]{>{\centering\arraybackslash}m{#1}}
    \begin{tabular}{M{0.4cm}M{4.1cm}M{4.1cm}M{4.1cm}M{4.1cm}}
    & \raisebox{-0.4\height}{(a) image id 0a4c0fa7a6fa9280} & \raisebox{-0.4\height}{(b) image id cb32cbcc6fb40d2a} & \raisebox{-0.4\height}{(c) image id 0a4aaf88891062bf} & \raisebox{-0.4\height}{(d) image id 4fb0e4b6908e0e56} \\
    \\
    \raisebox{-0.4\height}{\rotatebox[origin=c]{90}{Faster R-CNN}} & 
    \raisebox{-0.4\height}{\includegraphics[width=\linewidth]{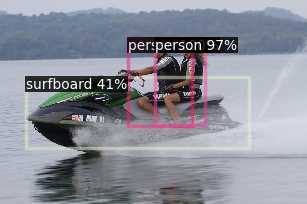}} & 
    \raisebox{-0.4\height}{\includegraphics[width=\linewidth]{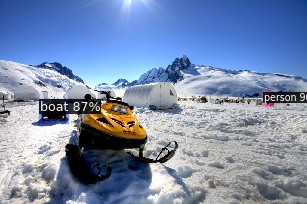}} & 
    \raisebox{-0.4\height}{\includegraphics[trim=0 0 0 15,clip,width=\linewidth]{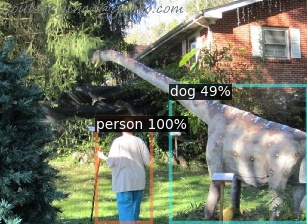}} &
    \raisebox{-0.4\height}{\includegraphics[trim=0 0 0 20,clip,width=\linewidth]{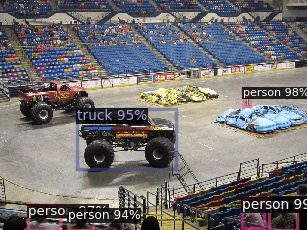}} \\
    \\
    \raisebox{-0.4\height}{\rotatebox[origin=c]{90}{Faster R-CNN KGE}} & 
    \raisebox{-0.4\height}{\includegraphics[width=\linewidth]{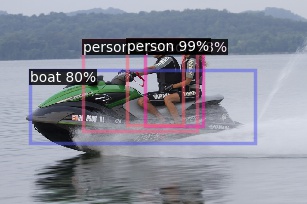}} &
    \raisebox{-0.4\height}{\includegraphics[width=\linewidth]{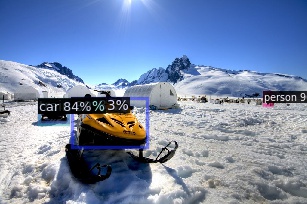}} &
    \raisebox{-0.4\height}{\includegraphics[trim=0 0 0 15,clip,width=\linewidth]{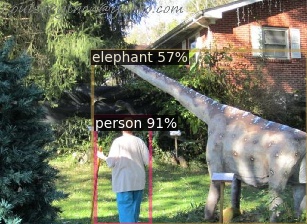}} &
    \raisebox{-0.4\height}{\includegraphics[trim=0 0 0 20,clip,width=\linewidth]{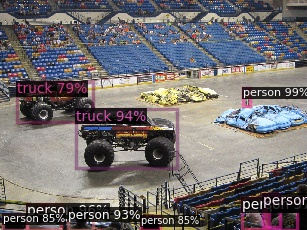}} \\
    \\
    \raisebox{-0.4\height}{\rotatebox[origin=c]{90}{CenterNet}} & 
    \raisebox{-0.4\height}{\includegraphics[width=\linewidth]{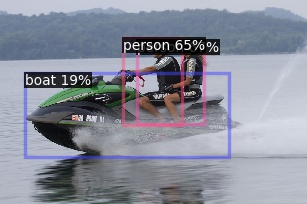}} & 
    \raisebox{-0.4\height}{\includegraphics[width=\linewidth]{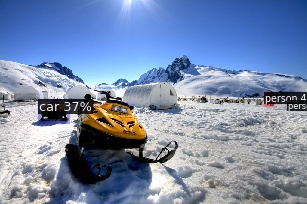}} & 
    \raisebox{-0.4\height}{\includegraphics[trim=0 0 0 15,clip,width=\linewidth]{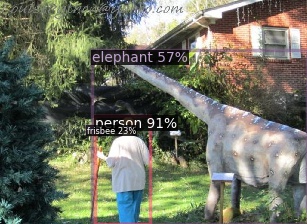}} &
    \raisebox{-0.4\height}{\includegraphics[trim=0 0 0 20,clip,width=\linewidth]{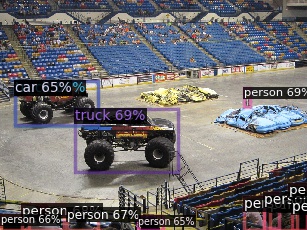}} \\
    \\
    \raisebox{-0.4\height}{\rotatebox[origin=c]{90}{CenterNet KGE}} & 
    \raisebox{-0.4\height}{\includegraphics[width=\linewidth]{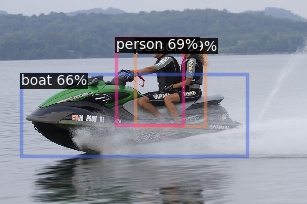}} & 
    \raisebox{-0.4\height}{\includegraphics[width=\linewidth]{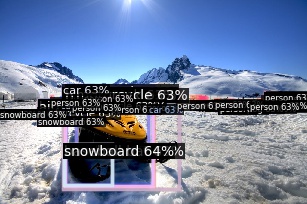}} & 
    \raisebox{-0.4\height}{\includegraphics[trim=0 0 0 15,clip,width=\linewidth]{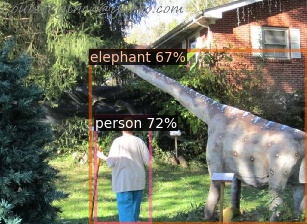}} &
    \raisebox{-0.4\height}{\includegraphics[trim=0 0 0 20,clip,width=\linewidth]{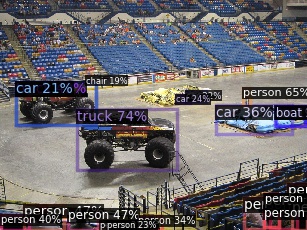}} \\
    \\
    \raisebox{-0.4\height}{\rotatebox[origin=c]{90}{DETR}} & 
    \raisebox{-0.4\height}{\includegraphics[width=\linewidth]{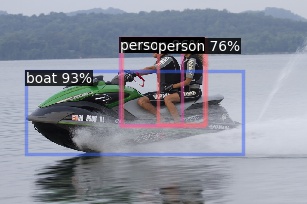}} & 
    \raisebox{-0.4\height}{\includegraphics[width=\linewidth]{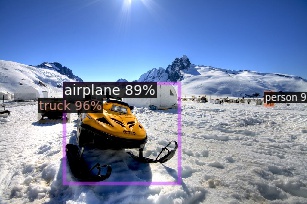}} & 
    \raisebox{-0.4\height}{\includegraphics[trim=0 0 0 15,clip,width=\linewidth]{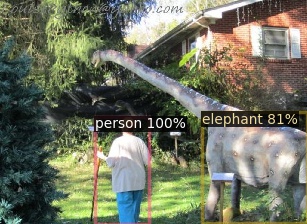}} &
    \raisebox{-0.4\height}{\includegraphics[trim=0 0 0 20,clip,width=\linewidth]{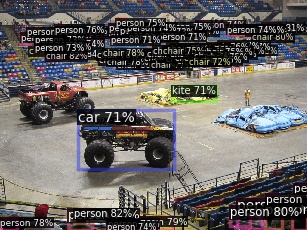}} \\
    \\
    \raisebox{-0.4\height}{\rotatebox[origin=c]{90}{DETR KGE}} & 
    \raisebox{-0.4\height}{\includegraphics[width=\linewidth]{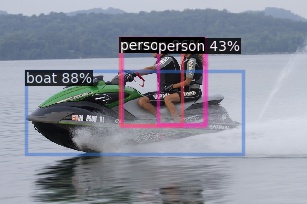}} & 
    \raisebox{-0.4\height}{\includegraphics[width=\linewidth]{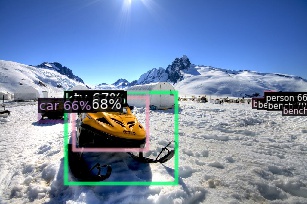}} & 
    \raisebox{-0.4\height}{\includegraphics[trim=0 0 0 15,clip,width=\linewidth]{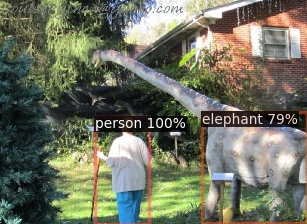}} &
    \raisebox{-0.4\height}{\includegraphics[trim=0 0 0 20,clip,width=\linewidth]{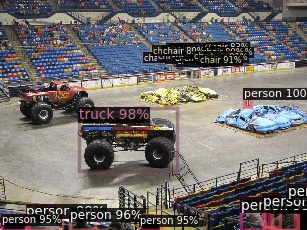}} \\
    \end{tabular}}
    \caption{Qualitative results on images from the  OpenImages ~\cite{OpenImages} validation set using the configurations in \tabref{tab:results_val_coco}. Please note that confidence scores for KGE methods describe the normalized similarity value between class prototype and visual embeddings.}
    \label{fig:qual-analysis}
\end{figure*}

We further qualitatively evaluate images from the OpenImages dataset~\cite{OpenImages} to demonstrate behavior on unseen objects and data distributions for methods trained on \acrshort{coco} dataset and classes. The results are shown in \figref{fig:qual-analysis}. All the methods demonstrate high object localization accuracy, where foreground region regression appears correlated with the choice of detection architecture. The \acrshort{kge}-based detection heads have consistently lower confidence on the detections, however, they skew the object classification to more semantically related classes, e.g. the unknown snow mobile in \figref{fig:qual-analysis}(b) is assigned to related classes such as \textit{car}, \textit{snowboard}, or \textit{motorcycle} by the \acrshort{kge} methods, rather than \textit{airplane}, \textit{boat}, or \textit{truck} by the baseline methods. The \acrshort{kge} methods also exhibit fewer false positives, such as the \textit{surfboard} for Faster R-CNN in \figref{fig:qual-analysis}(a) or \textit{kite} in \figref{fig:qual-analysis}(d) for DETR, which are not captured by the average precision metric on the object detection benchmarks.
    
    % the resulting class scores, such that they classify the velomobile in the middle frame as car (Faster R-CNN \acrshort{kge}) or motorcycle (CenterNet \acrshort{kge}), rather than boat (Faster R-CNN), frisbee (CenterNet) or airplane (DETR).

\section{Conclusions}
\label{sec:conclusions}
In this work, we demonstrate the transfer of feature embeddings from natural language processing or knowledge graphs as class prototypes into two-stage, keypoint-based, and transformer-based object detection architectures trained using a contrastive loss. We performed extensive ablation studies that analyze the choice of class prototypes and distance metrics on the \acrshort{coco} and Cityscapes datasets. We showed that our resulting method can compete with their standard configurations on challenging object detection benchmarks, with error distributions that are more consistent with the object categories in the groundtruth. Especially, class prototypes derived from the ConceptNet knowledge graph \cite{Speer2016} using a cosine distance metric demonstrate low inter-categorical confusions. Future work will investigate whether these knowledge embeddings in the classification head also benefit the class-incremental learning task in object detection.

% \section*{Acknowledgments}
% something

%%%%%%%%% REFERENCES
{\small
\bibliographystyle{IEEEtran}
\bibliography{references.bib}
}

%%%%%%%%%% Merge with supplemental materials %%%%%%%%%%
\newpage

\begin{strip}
\begin{center}
\vspace{-5ex}
\textbf{\Large \bf Contrastive Object Detection Using Knowledge Graph Embeddings} \\
\vspace*{12pt}

\Large{\bf Supplementary Material}\\
\vspace*{12pt}

\large{Christopher Lang$^{1,2}$ \hspace{1cm} Alexander Braun$^{1}$ \hspace{1cm} Abhinav Valada$^{2}$}\\
\large{$^{1}$Robert Bosch GmbH \hspace{1cm} $^{2}$University of Freiburg}\\
\vspace*{2pt}
\tt\small{\tt\small christopher.lang@de.bosch.com \hspace{1cm} valada@cs.uni-freiburg.de}

\end{center}
\end{strip}

%%%%%%%%%% Merge with supplemental materials %%%%%%%%%%
%%%%%%%%%% Prefix a "S" to all equations, figures, tables and reset the counter %%%%%%%%%%
\setcounter{section}{0}
\setcounter{equation}{0}
\setcounter{figure}{0}
\setcounter{table}{0}
\makeatletter

%\makeatletter \renewcommand{\fnum@figure}
%{\figurename~S\thefigure}
%\makeatother
% 
%% Hack for making figures Say \figurename S\thefigure, e.g. Figure S1:
%\makeatletter
%\makeatletter \renewcommand{\fnum@table}
%{\tablename~S\thetable}
%\makeatother

% Hack For section headers starting with S
%\renewcommand{\thesection}{S.\Roman{section}}
%\renewcommand{\thesubsection}{\thesection.\Alph{subsection}}
%\renewcommand{\bibnumfmt}[1]{[S#1]}
% citenumfont command adds S to all numbers
%\renewcommand{\citenumfont}[1]{\textit{S#1}}
%\renewcommand{\bibnumfmt}[1]{[S#1]}
%\renewcommand{\citenumfont}[1]{S#1}
%%%%%%%%%% Prefix a "S" to all equations, figures, tables and reset the counter %%%%%%%%%%

\normalsize

%\begin{figure*}
%    \centering
%    \footnotesize
%    \setlength{\tabcolsep}{0.5cm}% for the horiz padding
%    {\renewcommand{\arraystretch}{0.5}% for the vertical padding
%    \newcolumntype{M}[1]{>{\centering\arraybackslash}m{#1}}
%    \begin{tabular}{M{4.2cm}M{4.2cm}M{4.2cm}}
%    \includegraphics[width=1\linewidth]{figures/comparison/baseline1.png} & \includegraphics[width=1\linewidth]{figures/comparison/baseline2.png} &
%    \includegraphics[width=1\linewidth]{figures/comparison/baseline3.png} \\
%    \multicolumn{3}{c}{(a) AV-WaN~\cite{chen2020learning}} \\
%    \\
%    \includegraphics[width=1\linewidth]{figures/comparison/ours1.png} & 
%    \includegraphics[width=1\linewidth]{figures/comparison/ours2.png} &
%    \includegraphics[width=1\linewidth]{figures/comparison/ours3.png} \\
%    \multicolumn{3}{c}{(b) Ours}\\
%    \end{tabular}}
%    \captionof{figure}{Qualitative comparison of the dynamic audio-visual navigation task on the Replica dataset for heard sounds without complex scenarios. Each column shows the same episode for the AV-WaN agent on top and ours in the bottom. The paths of the agent and sound source are shown in blue and red, respectively. Start poses are marked with rectangles. Green shows the path to the earliest reachable intersection as defined for the DSPL metric.}
%    \label{fig:comparison}
%\end{figure*}
In this supplementary material, we provide additional insights and experimental results on knowledge embedding-based object detection.

%%%%%%%%% BODY TEXT
   
\section{Object Type Representation}

\begin{figure*}
        \centering
        \begin{subfigure}[b]{.9\columnwidth}
    		\centering
    		\includegraphics[width=\textwidth]{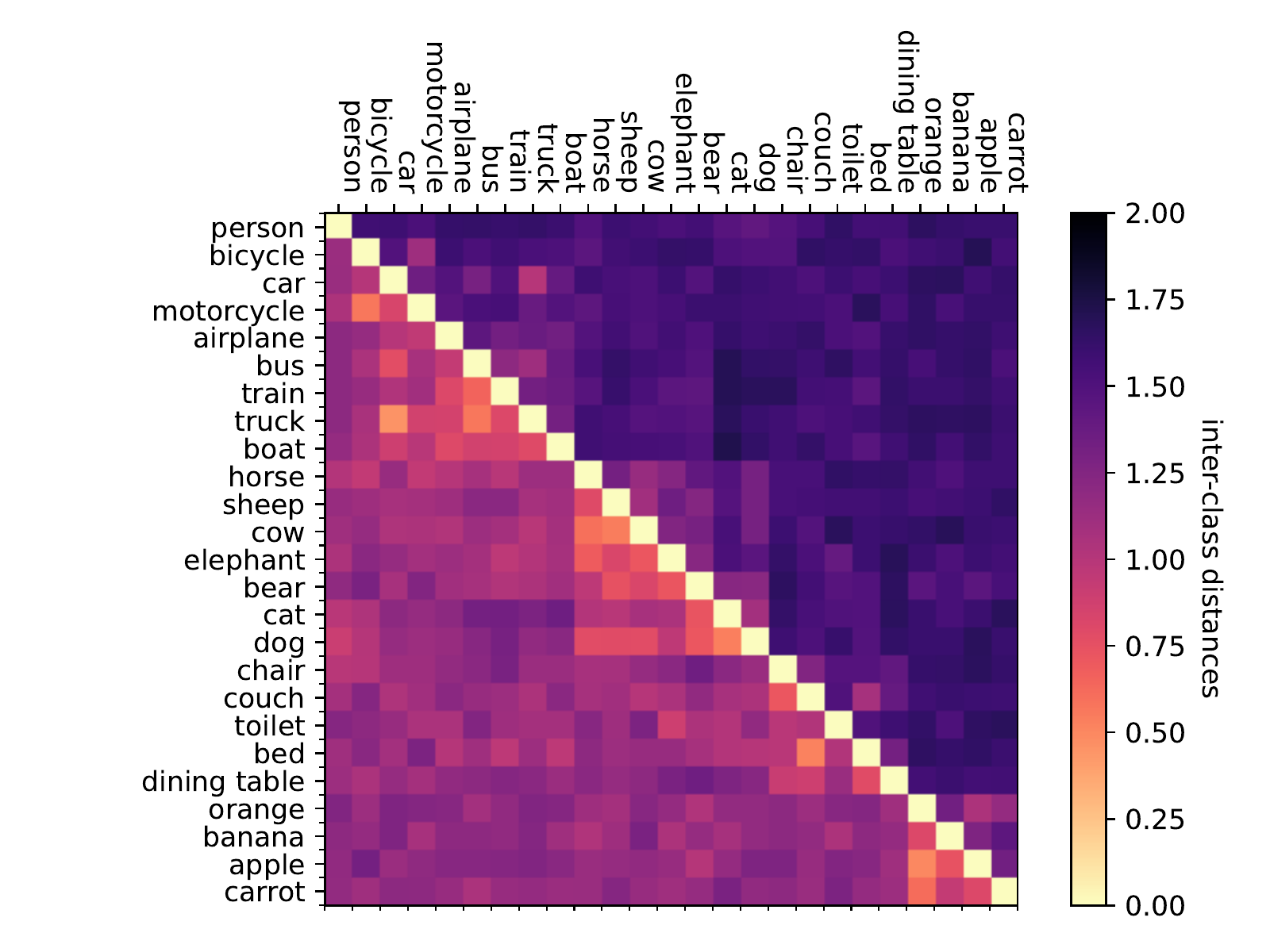}
    		\caption{Faster R-CNN trained embeddings, $D=1024$.}
    	\end{subfigure}%
        \begin{subfigure}[b]{.9\columnwidth}
    		\centering
    		\includegraphics[width=\textwidth]{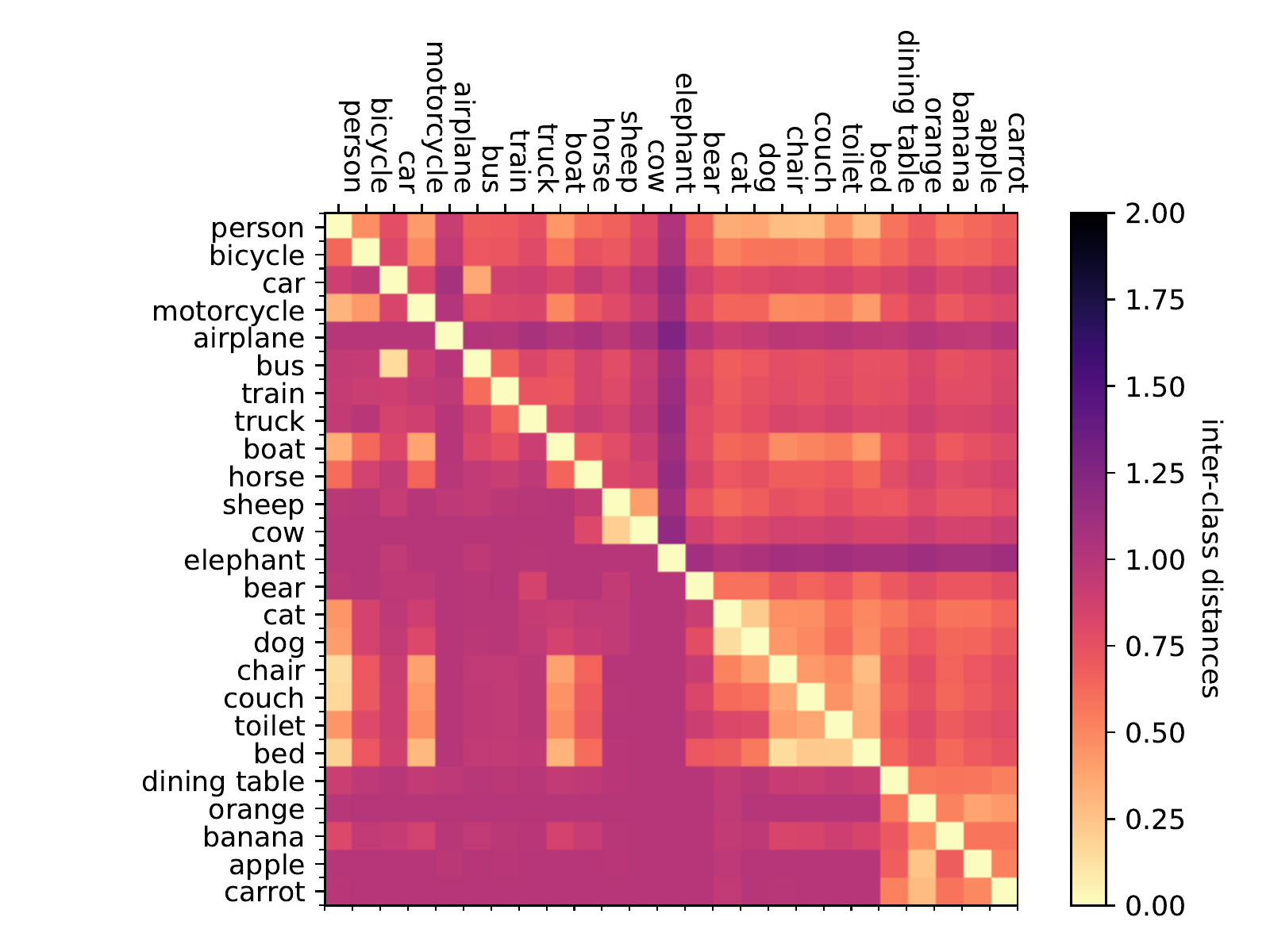}
    		\caption{COCO~\cite{Lin2014} spatial graph embeddings, $D=100$.}
    	\end{subfigure}%
    	
	    \begin{subfigure}[b]{.9\columnwidth}
    		\centering
    		\includegraphics[width=\textwidth]{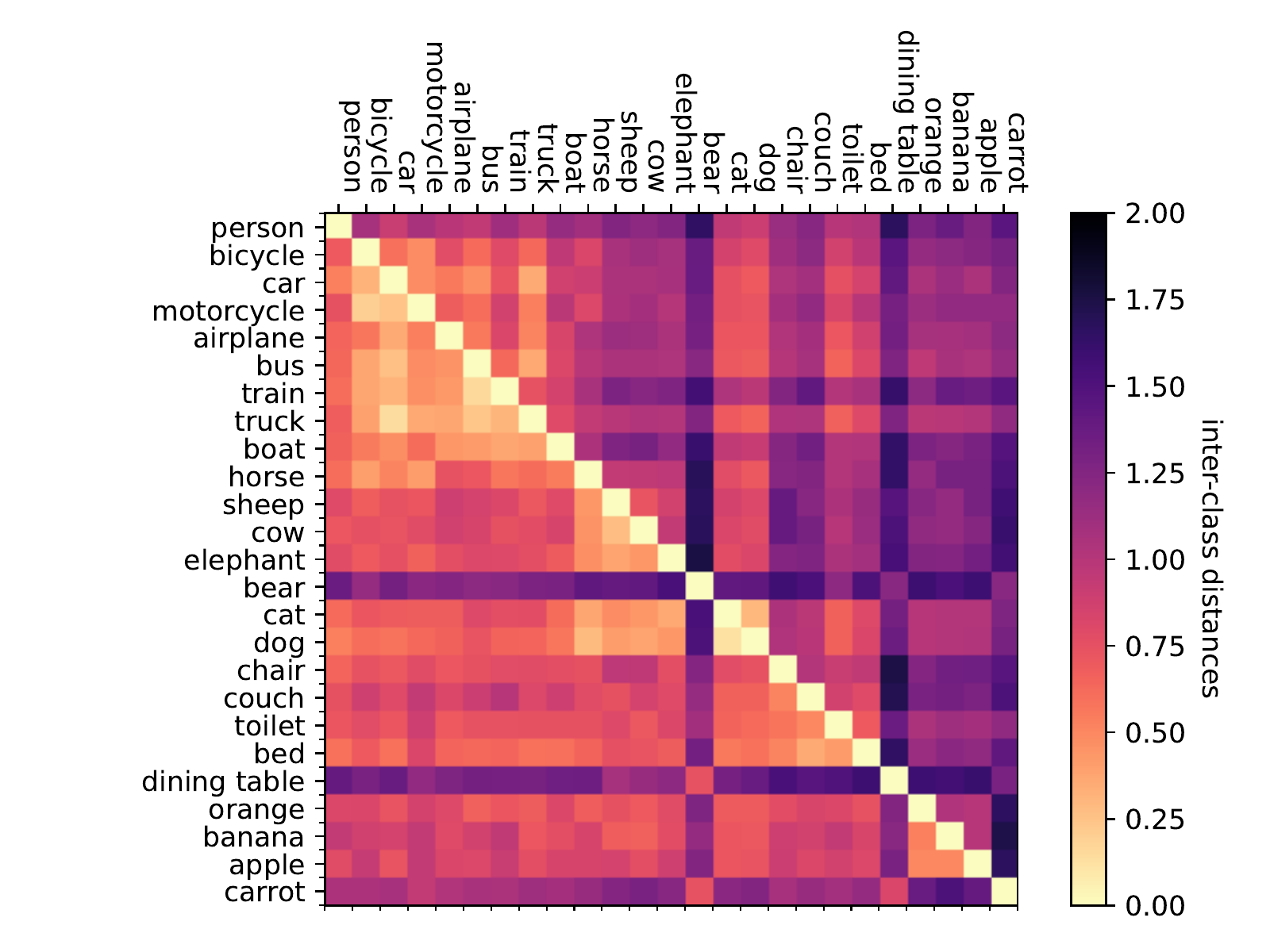}
    		\caption{GloVe word embeddings~\cite{glove}, $D=100$.}
    	\end{subfigure}%
	    \begin{subfigure}[b]{.9\columnwidth}
    		\centering
    		\includegraphics[width=\textwidth]{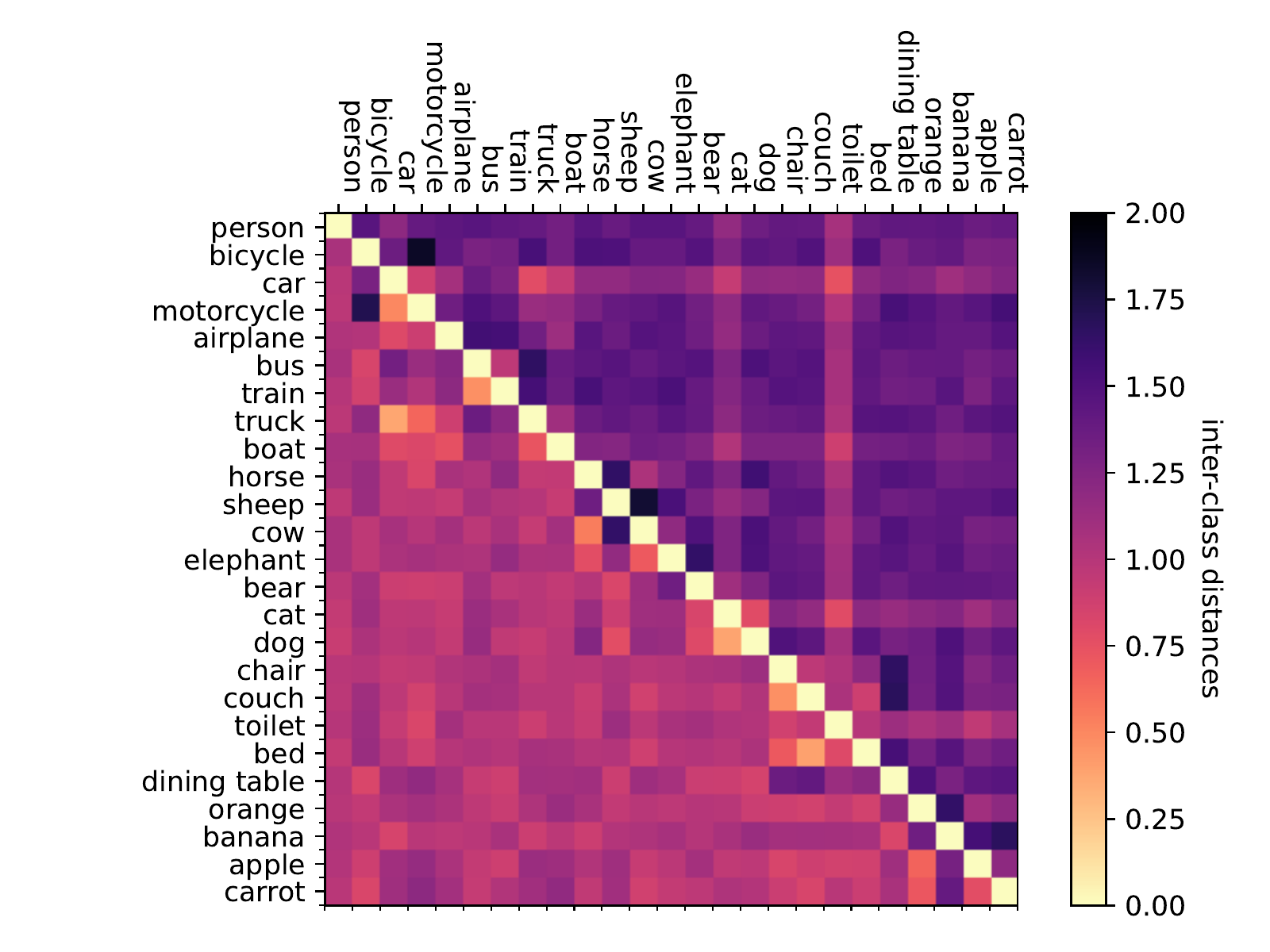}
    		\caption{ConceptNet graph embeddings~\cite{Speer2016}, $D=100$.}
    	\end{subfigure}%
        \caption{Pairwise distances matrix of class prototypes. Since the distance metrics are symmetric, we show cosine distances in the upper triangle and bounded Manhattan distances in the lower triangle.}
        \label{fig:pairwise_cossine}
    \end{figure*}

First, we inspect the knowledge embeddings that we incorporated as class prototypes into the object detection architectures.
\figref{fig:pairwise_cossine} summarizes the pairwise distances among a subset of the \acrshort{coco}~\cite{Lin2014} classes for each knowledge embedding. Please note that we exploit the symmetry of the distance metrics, and show cosine (upper triangle) and Manhattan distance (lower triangle) alongside in the same plot. The metrics share the main diagonal, as they are both zero for identical input vectors. The distances are bounded by $d(\mathbf{x}_1, \mathbf{x}_2) \in [0, 2]$, since the embedding vectors are normalized for the calculation of the cosine distance, respectively projected into the unit sphere for the calculation of the Manhattan distance.

In \figref{fig:pairwise_cossine} (a), we extracted the class prototypes from the penultimate layer of a Faster R-CNN~\cite{ren2015:faster-rcnn} model trained on the \acrshort{coco} dataset using a cross-entropy loss and learnable class prototypes. The method tends to maximize the pairwise distances among the class prototypes without semantic structure, except for weak clusters among groups of vehicle, animal, furniture, or food classes.

\begin{figure}
        \centering
        \includegraphics[trim=0 20 0 0, clip, width=\columnwidth]{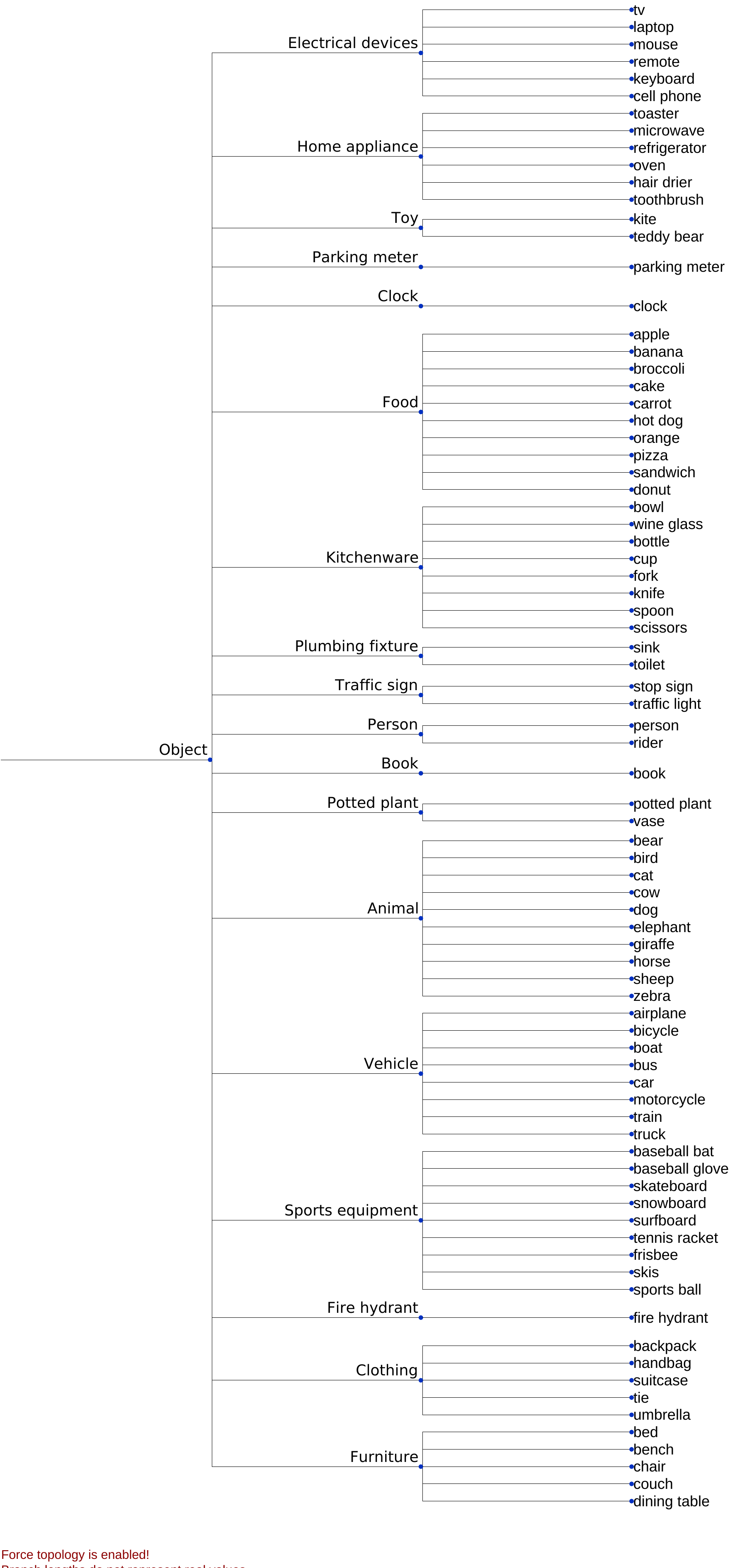}
        \caption{COCO class hierarchies extracted from the Google Knowledge Graph~\cite{google_knowledge_graph}. Categories are found from finding common parent nodes among all COCO 2017 \textit{thing} classes in the graph. The node identifiers are taken from the OpenImages \cite{OpenImages} class descriptions for matching object type labels.\looseness=-1}
        \label{fig:class_hierarchy}
    \end{figure}

\figref{fig:pairwise_cossine} (b) — (d) show the fixed class prototypes that were used during the experiments. Compared to the learned prototypes in (a), the distances are less uniform, reflecting the inherent semantic structure of the embeddings. The COCO embeddings (\figref{fig:pairwise_cossine} (b)) have noticeably low inter-class distances and depict clusters of commonly co-occurring objects such as vehicles, furniture, and pets, as well as foods. In the GloVe embedding~\cite{glove}~(\figref{fig:pairwise_cossine} (c)), we observe the most irregular distances end especially two outlier classes, such as bear or dining table, that have large distances to all other class embeddings. The ConceptNet embeddings~\cite{Speer2016} seem most similar to the learned embeddings with large inter-class distances, however, the inter-class distances show categorical differences, for example between furniture for sitting and lying, and vehicles for driving, swimming, or flying. The categorization of classes is shown in \figref{fig:class_hierarchy}. The categorization originates from traversing the Google knowledge graph \cite{google_knowledge_graph} starting from the \acrshort{coco} classes as leaf nodes and stopping at the first common parent node of each pair of classes.
\cite{Lin2014}

\begin{table}
\centering
\footnotesize
\begin{tabular}{ll|ccl}
\toprule
\multicolumn{2}{c|}{{ }} & \multicolumn{3}{c}{{COCO 2017 val}} \\
\rotatebox[origin=l]{90}{\thead{Distance metric}} & \rotatebox[origin=l]{90}{\thead{Background\\representation}} 
        & \multicolumn{1}{c}{{AP}} & \multicolumn{1}{c}{${AP_{w}}$} & \multicolumn{1}{c}{${AP_{cat,w}}$} \\ \midrule
Cossim   &  Explicit  & \textbf{39.4}	& 41.4	& 42.7  \\
Cossim   &  Implicit  & 39.3	& \textbf{41.6}	& \textbf{43.5}  \\
Manhattan   &  Explicit  & 36.0 &	38.1 & 38.9  \\
Manhattan   &  Implicit  & 37.0 &	39.3 & 40.5  \\
                                \bottomrule
\end{tabular}
\caption{Ablation study of Faster R-CNN \acrshort{kge} on the \acrshort{coco} validation set results over background representations using ConceptNet embeddings in 100 dimensions as class prototypes. Explicit background representations handle background regions as an additional class prototype vector for which nearest neighbor classification is performed, implicit background representations assign a background label based on a class confidence threshold.}
\label{tab:results_ablation_background}
\end{table}

\section{Extended Ablation Study}

In this section, we motivate the design choices we that make for the \acrshort{kge} object detection heads.
We therefore ran experiments on the Faster R-CNN~\cite{ren2015:faster-rcnn} base model using our proposed \acrshort{kge} extension as described in the main paper.

\subsection{Explicit vs Implicit background representation}
\label{sec:background}

The application of nearest neighbor classification brings the design choice of how to represent the background class.
\tabref{tab:results_ablation_background} shows our results when comparing an explicit background representation by an additional \textit{background} class prototype, or an implicit representation by a distance threshold to all foreground class representations. We found that an explicit representation results in marginally higher $AP$, however, this appears to be a misconception due to the non-weighted mean, while an implicit representation achieves higher $AP_w$ and $AP_{category,w}$ for the cosine distance metric. For Manhattan distances, the implicit representation surpasses the explicit representation on all metrics.

In our benchmark experiments, we choose an implicit background representation, motivated by the larger $AP_{category,w}$, which we interpret as an indicator of more semantically-grounded errors.
    
\subsection{Contrastive vs. Margin Loss}    

We experimented with both contrastive and margin embedding losses.
One representative of the latter is the Hinge embedding loss that, given a proposal embedding $\mathbf{b}_i$, minimizes the distance to the true class prototype while enforcing a fixed margin $\Delta$ to all other class prototypes as follows:
\begin{equation}
\resizebox{.9\hsize}{!}{$\mathcal{L}_d(\mathbf{b}_i) =
\frac{1}{C} \sum_{c=1}^{C}
\begin{cases}
d(\mathbf{b}_i, \mathbf{t}_c), & \text{ if } y_{i,c} = 1 \\ 
max\{0, \Delta - d(\mathbf{b}_i, \mathbf{t}_c)\}, & \text{ if } y_{i,c} = -1 
\end{cases} $}
\label{eq:hinge_embedding_loss}
\end{equation}
where $y_{i,c}$ is an indicator variable, that is $1$ if a proposal box $i$ is assigned a groundtruth bounding box of class $c$ by a matching algorithm (e.g., \acrshort{iou}-based or Hungarian matching), and $-1$ for all other classes. In our experiments, we set the distance margin to $\Delta=\max \{ 0.1, d(\mathbf{t}_{c}, \mathbf{t}_{f}) \}$, i.e., the distance between the true class embedding $\{\mathbf{t}_c|y_{i,c} = 1\}$ and all other class embeddings $\{\mathbf{t}_f|y_{i,f} = -1\}$ during training. Another hyperparameter is the negative sampling rate $r$ which denotes the number of randomly selected class prototypes other than the groundtruth class embedding evaluated for one proposal box embedding, for which we compute the margin loss term. For these experiments, we sample $r=5$ negative pairings uniformly from the set of \acrshort{coco} classes excluding the true class. As we can see from \tabref{tab:results_ablation_embedding}, the choice of embedding loss goes along with the distance metric that we use. A margin loss with a Manhattan distance metric achieves the highest $AP$ and $AP_w$. However, the contrastive loss performs better with a cosine distance metric and achieves the highest overall $AP_{category,w}$ with $43.5\%$.
We choose the contrastive loss as it spares us two hyperparameters, the fixed margin $\Delta$ and negative sampling rate $r$, with a negligible decline in average precision.

\begin{table}
\centering
\footnotesize
\begin{tabular}{ll|ccl}
\toprule
\multicolumn{2}{c|}{{ }} & \multicolumn{3}{c}{{COCO 2017 val}} \\
\rotatebox[origin=l]{90}{\thead{Distance metric}} & \rotatebox[origin=l]{90}{\thead{Embedding\\loss}} 
        & \multicolumn{1}{c}{{AP}} & \multicolumn{1}{c}{${AP_{w}}$} & \multicolumn{1}{c}{${AP_{cat,w}}$} \\ \midrule
Cossim   &  Margin  & 36.4	& 38.7	& 38.9  \\
Cossim   &  Contrastive  & 39.3	& 41.6	& \textbf{43.5}  \\
Manhattan   &  Margin  & \textbf{39.7} &	\textbf{41.8} & 43.0  \\
Manhattan   &  Contrastive  & 37.0 &	39.3 & 40.5  \\
                                \bottomrule
\end{tabular}
\caption{Ablation study of Faster R-CNN \acrshort{kge} on \acrshort{coco} validation set results over embedding loss configurations using ConceptNet embeddings in 100 dimensions as class prototypes.}
\label{tab:results_ablation_embedding}
\end{table}
\begin{table}
\centering
\footnotesize
\begin{tabular}{l|ccl}
\toprule
\multicolumn{1}{c|}{{ }} & \multicolumn{3}{c}{{COCO 2017 val}} \\
Contrastive loss $\tau$ 
        & \multicolumn{1}{c}{{AP}} & \multicolumn{1}{c}{${AP_{w}}$} & \multicolumn{1}{c}{${AP_{cat,w}}$} \\ \midrule
1.0  & 32.5 & 36.1 &	40.3  \\
0.1  & 33.0 & 36.1 &	40.1  \\
0.07 & \textbf{39.3} & \textbf{41.6} &	\textbf{43.5}  \\
                                \bottomrule
\end{tabular}
\caption{Ablation study of Faster R-CNN   \acrshort{kge} on \acrshort{coco} validation set results over temperature values for the contrastive loss using ConceptNet embeddings in 100 dimensions as class prototypes.}
\label{tab:results_ablation_temperature}
\end{table}
\begin{table}
\centering
\footnotesize
\begin{tabular}{lc|ccl}
\toprule
\multicolumn{2}{c|}{{ }} & \multicolumn{3}{c}{{COCO 2017 val}} \\
\rotatebox[origin=l]{90}{\thead{Distance metric}} & \rotatebox[origin=l]{90}{\thead{z-scores\\standardization}} 
        & \multicolumn{1}{c}{{AP}} & \multicolumn{1}{c}{${AP_{w}}$} & \multicolumn{1}{c}{${AP_{cat,w}}$} \\ \midrule
Cossim   &    & \textbf{39.3}	& \textbf{41.6}	& \textbf{43.5}  \\
Cossim   &  x  & 36.6	& 40.8	& 41.6  \\
Manhattan   &    & 37.0 &	39.3 & 40.5  \\
Manhattan   &  x  & 34.7 &	40.3 & 42.6  \\
                                \bottomrule
\end{tabular}
\caption{Ablation study of Faster R-CNN   \acrshort{kge} on \acrshort{coco} validation set results over standardization strategies using ConceptNet embeddings in 100 dimensions as class prototypes. }
\label{tab:results_ablation_standardization}
\end{table}

\subsection{High vs. Low Temperature Contrastive Losses}
\label{sec:temperature}

In the next step, we evaluated the inverse magnitude scaling parameter in the contrastive loss function, the so-called temperature $\tau$. Kornblith~\textit{et~al.}~\cite{Kornblith2020} noted that this parameter is a trade-off between generalizability and precision on the training data. \tabref{tab:results_ablation_temperature} shows the temperature parameter's effect on the validation set performance. High values for $\tau$ result in lower $AP$s since the distances are bounded by the interval $d \in [0,2]$, which results in flat scores after the approximate softmax scaling of the contrastive loss computation and thus low gradients. As we aim to achieve competitive performance as the baselines using cross-entropy losses, we follow \cite{Wang2020} and fix $\tau=0.07$.

\begin{figure*}
    \centering
    \footnotesize
    \setlength{\tabcolsep}{0.05cm}% for the horiz padding
    {\renewcommand{\arraystretch}{0.15}% for the vertical padding
    \newcolumntype{M}[1]{>{\centering\arraybackslash}m{#1}}
    \begin{tabular}{M{0.4cm}M{4.1cm}M{4.1cm}M{4.1cm}M{4.1cm}}
    & \raisebox{-0.4\height}{(a) image id 0a1dae152634618c} & \raisebox{-0.4\height}{(b) image id fd81aed21b5d54b8} & \raisebox{-0.4\height}{(c) image id 3a0b72a274530636} & \raisebox{-0.4\height}{(d) image id 040ad403d9c14ee2} \\
    \\
    \raisebox{-0.4\height}{\rotatebox[origin=c]{90}{Faster R-CNN}} & 
    \raisebox{-0.4\height}{\includegraphics[width=\linewidth]{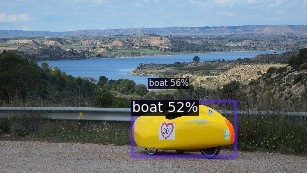}} & 
    \raisebox{-0.4\height}{\includegraphics[width=\linewidth]{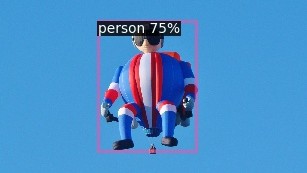}} & 
    \raisebox{-0.4\height}{\includegraphics[width=\linewidth]{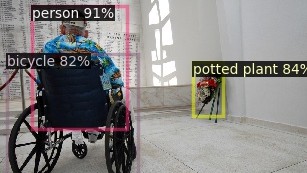}} &
    \raisebox{-0.4\height}{\includegraphics[width=\linewidth]{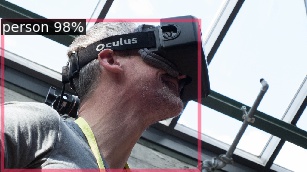}} \\
    \\
    \raisebox{-0.4\height}{\rotatebox[origin=c]{90}{Faster R-CNN KGE}} & 
    \raisebox{-0.4\height}{\includegraphics[width=\linewidth]{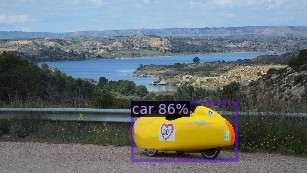}} &
    \raisebox{-0.4\height}{\includegraphics[width=\linewidth]{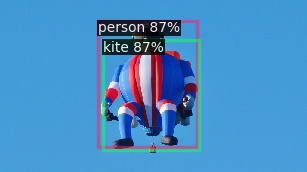}} &
    \raisebox{-0.4\height}{\includegraphics[width=\linewidth]{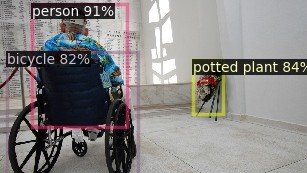}} &
    \raisebox{-0.4\height}{\includegraphics[width=\linewidth]{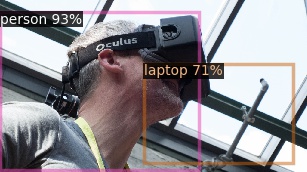}} \\
    \\
    \raisebox{-0.4\height}{\rotatebox[origin=c]{90}{CenterNet}} & 
    \raisebox{-0.4\height}{\includegraphics[width=\linewidth]{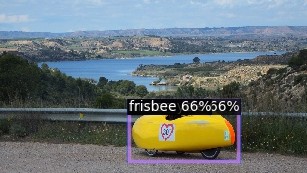}} & 
    \raisebox{-0.4\height}{\includegraphics[width=\linewidth]{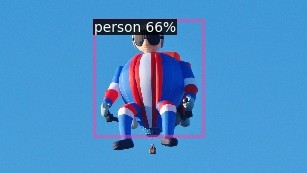}} & 
    \raisebox{-0.4\height}{\includegraphics[width=\linewidth]{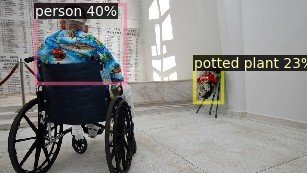}} &
    \raisebox{-0.4\height}{\includegraphics[width=\linewidth]{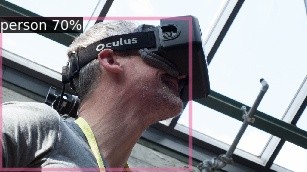}} \\
    \\
    \raisebox{-0.4\height}{\rotatebox[origin=c]{90}{CenterNet KGE}} & 
    \raisebox{-0.4\height}{\includegraphics[width=\linewidth]{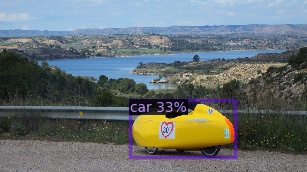}} & 
    \raisebox{-0.4\height}{\includegraphics[width=\linewidth]{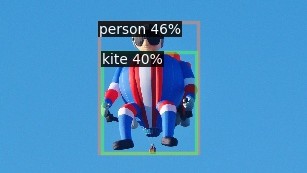}} & 
    \raisebox{-0.4\height}{\includegraphics[width=\linewidth]{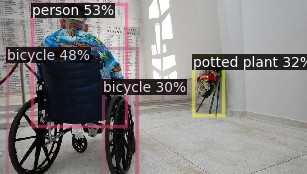}} &
    \raisebox{-0.4\height}{\includegraphics[width=\linewidth]{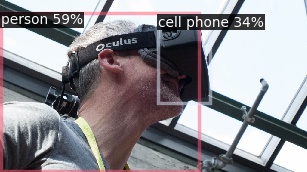}} \\
    \\
    \raisebox{-0.4\height}{\rotatebox[origin=c]{90}{DETR}} & 
    \raisebox{-0.4\height}{\includegraphics[width=\linewidth]{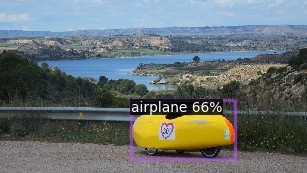}} & 
    \raisebox{-0.4\height}{\includegraphics[width=\linewidth]{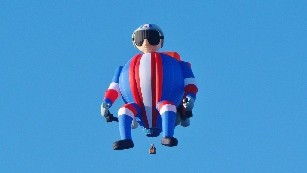}} & 
    \raisebox{-0.4\height}{\includegraphics[width=\linewidth]{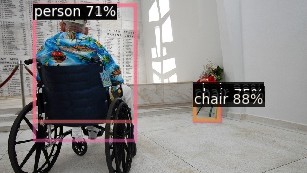}} &
    \raisebox{-0.4\height}{\includegraphics[width=\linewidth]{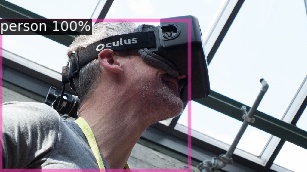}} \\
    \\
    \raisebox{-0.4\height}{\rotatebox[origin=c]{90}{DETR KGE}} & 
    \raisebox{-0.4\height}{\includegraphics[width=\linewidth]{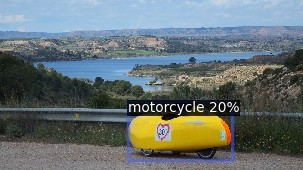}} & 
    \raisebox{-0.4\height}{\includegraphics[width=\linewidth]{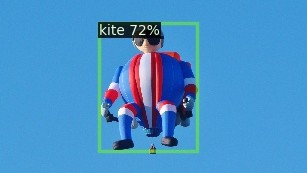}} & 
    \raisebox{-0.4\height}{\includegraphics[width=\linewidth]{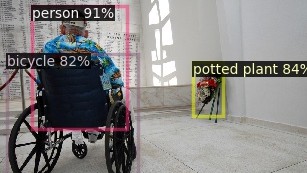}} &
    \raisebox{-0.4\height}{\includegraphics[width=\linewidth]{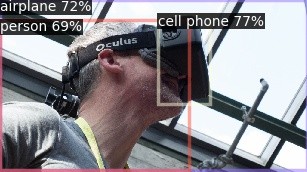}} \\
    \end{tabular}}
    \caption{Qualitative results on images from the OpenImages~\cite{OpenImages} validation set using \acrshort{kge} configurations with cosine distance and ConceptNet \cite{Speer2016} class prototypes. Please note that confidence scores for KGE methods describe the normalized similarity value between class prototype and visual embeddings.}
    \label{fig:qual-analysis-supp}
\end{figure*}

\subsection{Hubness Reduction vs. No Hubness Reduction}
\label{sec:zscore}

Fei~\textit{et~al.}~\cite{fei2021z} recently raised awareness for the hubness problem in image classification. The hubness problem is a phenomenon that concerns nearest neighbor classification in high dimensional spaces. It relates to the distance concentration effect and results in few class prototypes being particularly often among the nearest neighbors of embeddings, while others are almost never. They report that local feature scaling by z-score standardization can reduce hubness in image classification. In this work we aim to validate whether z-score standardization can improve the classification accuracy as indirectly measured by the $AP$.

The results presented in \tabref{tab:results_ablation_standardization} shows that a hubness reduction measure such as z-score standardization does not improve $AP$ for neither cosine nor Manhattan distance. On the contrary, it hinders convergence of the training, such that the resulting validation set performance is worse when using z-score normalization. Therefore, we conclude that the fixed knowledge embeddings already bring sufficient training data and internal structure to alleviate hubs in the feature space.

\section{Extended Qualitative Results}

In this section, we discuss the out-of-distribution predictions on the OpenImages~\cite{OpenImages} dataset shown in \figref{fig:qual-analysis-supp}, for models trained on \acrshort{coco} data and classes. In general, we observe that the object detectors mostly detect and localize all foreground objects in the frame, even if they are from unknown classes.  The \acrshort{kge} modification appears to predominantly impact the classification scores and results in more semantically-grounded class predictions. For instance, the velomobile in \figref{fig:qual-analysis-supp}~(a) is classified as car or motorcycle by the \acrshort{kge} models, rather than boat, Frisbee, or airplane as observed in the predictions from the standard methods. The same holds for the pilot-shaped hot air balloon in \figref{fig:qual-analysis-supp}~(b), which is consistently identified as a kite using the knowledge-embedded class prototypes.

The \acrshort{kge} methods also predict reasonable high confidences to more abstract objects such as the wheelchair or VR goggles in \figref{fig:qual-analysis-supp}~(c) or \figref{fig:qual-analysis-supp}~(d), while the standard models classify such objects as background. The \acrshort{kge} methods produce overall lower confidence scores compared to their non-\acrshort{kge} counterparts, however, this is due to the variable inter-class distances in the knowledge embeddings.

\end{document}